\documentclass{article} %
\usepackage{iclr2027_conference,times}

\usepackage{latexsym}

\usepackage[T1]{fontenc}

\usepackage[utf8]{inputenc}

\usepackage{microtype}

\usepackage{inconsolata}

\usepackage{graphicx}

\usepackage{xcolor} %
\usepackage{arydshln}
\usepackage{ulem}
\usepackage{booktabs}
\usepackage{array}
\usepackage{tabularx}
\usepackage{float}
\usepackage{listings}
\lstdefinestyle{prompt}{
  basicstyle=\ttfamily\footnotesize,
  breaklines=true,
  breakindent=0pt,
  columns=fullflexible,
  keepspaces=true,
  showstringspaces=false,
  aboveskip=0.6em,
  belowskip=0.6em
}
\usepackage{multirow}
\usepackage{pifont}
\usepackage{hyperref}
\usepackage{url}

\newcommand{\cmark}{\ding{51}}  % Checkmark
\newcommand{\xmark}{\ding{55}}  % X mark
\definecolor{darkgreen}{rgb}{0.2, 0.46, 0.05}

\newenvironment{transcript}
  {\par\smallskip\small
   \renewcommand{\texttt}[1]{{\upshape\ttfamily ##1}}%
   \begin{list}{}{%
     \setlength{\leftmargin}{3.6em}%
     \setlength{\labelwidth}{3.2em}%
     \setlength{\labelsep}{0.4em}%
     \setlength{\itemsep}{1.5pt}%
     \setlength{\parsep}{0pt}%
     \setlength{\topsep}{1pt}%
     \setlength{\rightmargin}{0.5em}%
     \raggedright
     \hyphenpenalty=10000\relax
     \exhyphenpenalty=10000\relax
     \renewcommand{\_}{\textunderscore\allowbreak}}}
  {\end{list}\smallskip}
\newcommand{\tuser}{\item[\textsc{user}]}
\newcommand{\tagent}{\item[\textsc{agent}]}
\newcommand{\tcall}[1]{\item[\textsc{call}]{\upshape\ttfamily\footnotesize #1}}
\newcommand{\ttool}[1]{\item[\textsc{tool}]{\footnotesize #1}}
\newcommand{\tjudge}[1]{\item[\textsc{judge}]{\itshape #1}}
\newcommand{\tomit}[1]{\item[]{\itshape [#1]}}

\newcommand{\myemph}[1]{\textbf{#1}} %

\newcommand{\myparagraph}[1]{\textbf{#1}}

\title{ToolTurns: Benchmarking LLM Agents in Multi-Turn Dialogue}

\author{
Milad Gholami \qquad Brendan King \qquad Jeffrey Flanigan\\
University of California, Santa Cruz\\
\texttt{\{mgholami,bking2,jmflanig\}@ucsc.edu}
}

\hypersetup{
  pdftitle={ToolTurns: Benchmarking LLM Agents in Multi-Turn Dialogue},
  pdfauthor={Milad Gholami, Brendan King, Jeffrey Flanigan}
}
\iclrfinalcopy
\begin{document}
\maketitle
\lhead{Preprint}
\begin{abstract}
Tool-using language models are increasingly deployed in conversational settings, yet most large-scale tool-use benchmarks focus on isolated requests. We introduce ToolTurns, a benchmark of 2,620 problems spanning 1,827 tools and 11,004 API endpoints across 47 categories. Each dialogue begins with an intentionally underspecified request and continues through two context-dependent follow-ups, testing clarification, tool use, and retention of conversational context. We also introduce an open-source user simulator that reveals withheld context through dialogue and generates follow-up requests, with a checker--fixer pipeline for improving its turns. Across nine language models, the best agent resolves 69.6\% of individual requests but completes only 39.8\% of full dialogues. Under this evaluation protocol, every evaluated model receives lower success scores on follow-up requests than on the initial request, and dialogue-quality rubric scores and task success can rank models very differently. Paired turn-level evaluation shows that the checker--fixer reduces detected dialogue defects, and blind human evaluation favors the revised turns. ToolTurns provides a testbed for evaluating whether agents sustain task success across multi-turn dialogue with tools.
\end{abstract}

\section{Introduction}

A tool-using assistant must do more than answer an initial request.
A user asking ``What games are happening this weekend?'' may omit the sport or location, then ask follow-up questions about the results.
The assistant must elicit missing information, use the available APIs, and carry relevant context into subsequent requests.
These decisions shape the conversation itself: an assistant's clarification or answer changes what the user says next.
Evaluating this behavior therefore requires following the agent through an interaction, including the requests that arise after its first answer.

Existing benchmarks capture different aspects of this challenge (\autoref{tab:related-work}).
ToolBench \citep{qin2024toolllm} provides broad API coverage, while ToolDial \citep{shim_tooldial_2025} evaluates conversational tool-use capabilities from recorded dialogue histories.
Interactive benchmarks such as $\tau$-bench \citep{yao_-bench_2024} and ToolSandbox \citep{lu-etal-2025-toolsandbox} evaluate agents within stateful application environments; $\tau^{2}$-bench \citep{barres_tau2_2025} further tests coordination when users can also act on the environment.
We study the combination of broad API coverage and adaptive, end-to-end dialogue evaluation: whether agents can clarify underspecified requests and resolve related follow-ups across diverse tool sets.

We introduce ToolTurns, a benchmark of 2,620 initially underspecified problems spanning 1,827 tools and 11,004 API endpoints across 47 RapidAPI categories (\autoref{sec:benchmark}; statistics in \autoref{tab:benchmark-stats}).
Each problem exposes a fixed subset of endpoints---26.5 on average---and unfolds into an initial request followed by two adaptively generated follow-up requests.
A user simulator answers clarification questions using withheld context and generates follow-ups from the user-visible conversation.
Agents interact with these users and a StableToolBench-based tool environment \citep{guo-etal-2024-stabletoolbench} (a cached, LLM-simulated execution layer for the ToolBench APIs), producing 7,860 separately scored subtasks per agent.
This design combines API breadth with evaluation along each agent's own conversational trajectory.

Because the simulated user helps determine the evaluation trajectory, we also examine the quality of its contributions.
We develop a checker--fixer pipeline that revises candidate user messages before they enter the conversation (\autoref{sec:benchmark-user-sim}).
In a paired analysis, revision reduces rule-detected defects by 41.2\% among the messages it rewrites (536 of 1{,}272 candidates); blind human annotation prefers the revised message in 63.3\% of 120 sampled pairs (Appendix~\ref{sec:exp-revision-turnlevel}, Appendix~\ref{sec:exp-human-eval}).
We evaluate task completion separately from dialogue quality, reporting both the fraction of requests resolved and the fraction of dialogues in which all three requests succeed.

Across nine proprietary and open-weight agent models, our evaluator reports a highest subtask success rate of 69.6\% and a highest full-dialogue completion rate of 39.8\% (\autoref{sec:experiments}).
Under the existing evaluation protocol, every evaluated model receives lower success scores on both follow-ups than on the initial request.
Task-success rankings also differ from rankings under the dialogue-quality rubric.
These findings highlight the importance of evaluating completion throughout a conversation and measuring it directly alongside diagnostic quality scores.

\noindent \textbf{Our contributions are:}
\begingroup
\setlength{\topsep}{1pt}%
\setlength{\partopsep}{0pt}%
\setlength{\parsep}{0pt}%
\setlength{\itemsep}{1pt}%
\setlength{\parskip}{0pt}%
\begin{enumerate}
\item ToolTurns, a benchmark combining broad API coverage with adaptive dialogue: 2,620 problems with clarification exchanges and follow-up requests evaluated along the agent's own trajectory.
\item A user-simulation pipeline with targeted message revision, evaluated through paired automatic measurements and blind human annotation.
\item An evaluation of nine agents that characterizes request-level and full-dialogue completion, follow-up performance, and differences between task-success and dialogue-quality rankings.
\end{enumerate}
\endgroup

\section{Related Work}
\label{sec:related-work}

The rise of tool-calling LLMs has led to two key directions in evaluation which intersect our work: scaling to large, diverse sets of APIs, and supporting tool use in interactive, multi-turn dialogues.

\myparagraph{Scaling Tool Evaluation} Where early tool-calling evaluations focused on relatively small sets of tools \citep{tang_toolalpaca_2023}, recent efforts have aimed to scale this to larger, more diverse sets of tools \citep{patil_gorilla_2023, qin2024toolllm, elder_live_2026}.
\citet{qin2024toolllm} provide one of the largest such evaluations in ToolBench, evaluating LLMs on problems using over 16,000 APIs sourced from RapidAPI.
\footnote{\url{https://rapidapi.com/}}
While comprehensive in tool scope, each problem is designed for only a single-turn interaction: missing real-world cases where the model must seek details or clarification from the user, or respond to follow-ups.
Our work aims to address this limitation by introducing ToolTurns, a large-scale tool-calling evaluation for the multi-turn dialogue setting, where an agent must support both clarification-seeking and follow-up requests.
We build our evaluation upon the RapidAPI tool set proposed in \citet{qin2024toolllm}.

\myparagraph{Interactive Tool-Use Benchmarks}
Prior tool-use evaluations fall into three settings: broad API corpora, recorded conversational evaluations, and interactive stateful environments (\autoref{tab:related-work}).
ToolBench gives one user instruction over a large RapidAPI catalogue, so later tool steps are not further user turns; ToolDial, ToolTalk, API-Bank, and DialogTool instead score recorded or annotated dialogues, executing tools only where the benchmark implements them \citep{qin2024toolllm, shim_tooldial_2025, farn2023tooltalkevaluatingtoolusageconversational, li-etal-2023-api, wang-etal-2025-dialogtool}.
$\tau$-bench, $\tau^{2}$-bench, ToolSandbox, and VitaBench keep an adaptive user inside a stateful application and grade environment outcomes or rubrics \citep{yao_-bench_2024, barres_tau2_2025, lu-etal-2025-toolsandbox, he2026vitabench}.

ToolTurns evaluates clarification and successive requests over a broad API catalogue.
$\tau$-bench and $\tau^{2}$-bench emphasize domain policies, verifiable environment outcomes, and, for $\tau^{2}$ telecom, user--agent coordination.
Our broader schema coverage does not establish stronger state modeling or more reliable ground-truth evaluation.
Two related settings sit outside the table.
\citet{wang2024mint} evaluate multi-turn tool use in which the model executes Python tools and revises its solution using language feedback.
\citet{wu-etal-2024-toolplanner} study tool planning for multi-granularity instructions; those solution steps are successive tool calls for one instruction, not conversational turns.

\begin{table}[H]
\centering
\small
\setlength{\tabcolsep}{2.5pt}
\renewcommand{\arraystretch}{1.08}
\begin{tabularx}{\textwidth}{@{}>{\raggedright\arraybackslash}p{0.82in} >{\raggedright\arraybackslash}X >{\raggedright\arraybackslash}X >{\raggedright\arraybackslash}X >{\raggedright\arraybackslash}X@{}}
\toprule
\textbf{Benchmark} & \textbf{Resource size} & \textbf{Tool/API scope} & \textbf{User interaction during evaluation} & \textbf{Evaluation target} \\
\midrule
ToolBench & 126{,}486 instruction--solution pairs (corpus) & 3{,}451 tools; 16{,}464 APIs; 49 categories & Single user instruction; tool steps are not user turns & Pass rate and win rate (ToolEval) \\
\addlinespace
ToolDial & 11{,}111 generated dialogues (corpus) & RapidAPI graph (4{,}474 APIs) & Recorded dialogue histories & Action, API-argument, and response prediction \\
\addlinespace
ToolTalk & 78 annotated conversations & 28 callable tools & Fixed user turns; predicted calls executed with feedback & Conversation success from tool-call match \\
\addlinespace
API-Bank & 314 annotated dialogues (evaluation) & 73 runnable tools & Annotated dialogue context & Planning, retrieval, and calls \\
\addlinespace
$\tau$-bench & 115 retail tasks; 50 airline tasks & 15 retail and 13 airline API tools & Adaptive simulated user & Database outcome and repeated-trial reliability \\
\addlinespace
$\tau^{2}$-bench & 115 retail; 50 airline; 114 telecom tasks & Agent tools: 13 retail, 12 airline, 13 telecom; 30 telecom user tools & Adaptive user; dual control in telecom & Verifiable task outcomes \\
\addlinespace
ToolSandbox & 1{,}032 evaluation scenarios & 34 tools & Adaptive simulated user; stateful execution & Milestone and minefield checks \\
\addlinespace
VitaBench & 400 tasks (100 cross-scenario; 300 single-scenario) & 66 tools & Interactive application environments & Rubric-based outcomes \\
\addlinespace
DialogTool & 900 evaluation dialogues & 15 apps; 30 APIs (evaluation) & Fixed dialogue context; APIs execute in VirtualMobile & Creation, selection, execution, and response scores \\
\addlinespace
ToolTurns & 2{,}620 problems; 3 scored requests each & 1{,}827 tools; 11{,}004 endpoints; 47 categories & Adaptive clarification and follow-up requests & Subtask and full-dialogue completion; dialogue-quality diagnostics (LLM judge) \\
\bottomrule
\end{tabularx}
\caption{How the evaluation setting differs across benchmarks.
ToolTurns combines broad API coverage with adaptive clarification and follow-up dialogue, complementing benchmarks centered on stateful application workflows.
Counting conventions for each row are in Appendix~\ref{sec:related-work-notes}.}
\label{tab:related-work}
\end{table}

\section{Benchmark Design}
\label{sec:benchmark}

\subsection{Problem Design}
\label{sec:bench-problem-design}

A ToolTurns problem is a triple $(q_i, T_i, G_i)$, where $q_i$ is the user's initial natural-language query, $T_i$ is a fixed set of callable tool endpoints and their schemas, and $G_i = \{(s_{ij}, v_{ij})\}_{j=1}^{m_i}$ is the unseen user's goal, represented as a set of slot--value constraints.
Each slot $s_{ij}$ corresponds to an argument in a tool schema, and $v_{ij}$ is the value that the user has in mind but does not initially specify in $q_i$.

The dialogue begins with $q_i$.
At each step $t$, conditioned on the dialogue history and the schemas of $T_i$, the agent selects an action $a_t \in \mathcal{A}(T_i)$ that either (1) invokes an endpoint and receives a tool observation $o_t$, (2) sends a message to the user and receives a user response $o_t$, or (3) completes the current request with a final answer $y_t$.
In response to a clarification request, the user simulator reveals relevant slot values from $G_i$.

This design isolates two distinct skills: \myemph{tool selection and use} (the agent must combine endpoints from the available set to satisfy a request) and \myemph{clarification dialogue} (the agent must elicit the withheld context before it can call the right endpoint with the right parameters).

\myparagraph{Tool catalogue.}
We use the RapidAPI tool collection released with ToolBench \citep{qin2024toolllm}, in which each tool is represented by its available endpoints and parameter schemas.
StableToolBench exposes the same collection during execution, allowing the sampled tool sets to be evaluated without schema conversion.

\myparagraph{Tool-set sampling.}
We draw seed tool sets from ToolBench's three retrieval groups, which differ in how the underlying tools relate to one another:
\myemph{G1} (single-tool sets), \myemph{G2} (intra-category multi-tool sets, where the tools share a RapidAPI category), and \myemph{G3} (intra-collection multi-tool sets, which span categories within a curated collection).
Following ToolBench's own filtering, we discard seed sets with fewer than three endpoints.
G1 always uses a single seed set per problem.
For G2 and G3, with probability 0.5 we sample two distinct seed sets and combine their endpoints; this produces problems that may require tools from different categories.

\myparagraph{Endpoint expansion.}
ToolBench's seed sets identify a small number of representative endpoints, but realistic requests may require other capabilities of the same tools.
We therefore expand each sampled set to include all endpoints belonging to its referenced tools.

\myparagraph{Vague-query generation.}
For each expanded tool set, we prompt GPT-4o-mini to produce two queries using a fixed template. 
The prompt instructs the model that queries must (a)~read like a natural request a human might make, (b)~not mention tool or API names, (c)~be answerable by some combination of the provided endpoints, and (d)~deliberately omit details that the agent would need to actually make a call.
For each query the model additionally returns 2--4 \myemph{context keywords} encoding the omitted details as concrete values (e.g.\ \texttt{Paris}, \texttt{December}, \texttt{Lakers}) rather than type placeholders (e.g.\ \texttt{city}, \texttt{month}, \texttt{team\_name}); the prompt gives explicit good/bad examples to enforce this.
These keywords become the problem's withheld context.

\myparagraph{Example.}
Problem 2610 in ToolTurns has the initial query:
\begin{quote}
\small
\textit{``I heard there's a big game coming up soon. Can you tell me when it is?''}
\end{quote}
Its withheld context is \texttt{NBA}, \texttt{Los Angeles}, \texttt{Lakers}, and \texttt{October}, and its tool set exposes 48 endpoints from an NBA data API, a web search API, and a geographic database.
The agent must first clarify the sport, team, and time frame before selecting among the available endpoints.
\autoref{sec:failure-example-running} shows how different agents handle this problem.

\begin{table}[t]
\centering
\small
\begin{tabular}{lr}
\toprule
\textbf{Metric} & \textbf{Value} \\
\midrule
Total problems & 2{,}620 \\
\quad Single-tool & 1{,}032\ (39.4\%) \\
\quad Multi-tool (2+) & 1{,}588\ (60.6\%) \\
\quad\quad Cross-category & 1{,}023\ (64.4\% of multi) \\
\midrule
Distinct tools & 1{,}827 \\
Distinct API endpoints & 11{,}004 \\
RapidAPI categories & 47 \\
\midrule
Tools / problem & 3.4 / 3.0 \\
Endpoints / problem & 26.5 / 15.0 \\
\bottomrule
\end{tabular}
\caption{ToolTurns statistics for the 2,620-problem evaluation set. The last two rows report mean\,/\,median.}
\label{tab:benchmark-stats}
\end{table}

\subsection{Dialogue Generation Protocol}
\label{sec:bench-dialogue-protocol}

A dialogue is generated by alternating between an \myemph{agent} (the assistant under evaluation), a \myemph{user simulator} (\autoref{sec:benchmark-user-sim}), and the \myemph{tool environment} provided by StableToolBench.
For each problem, the environment exposes only the sampled set of endpoints.
Each problem (\autoref{sec:bench-problem-design}) yields exactly one dialogue.

\myparagraph{Agent turn structure.}
Within a single turn the agent runs a tool-using search loop.
Two reserved control tools terminate the turn: \texttt{finish}, which the agent calls when it has a final answer for the user, and \texttt{message\_user}, which the agent calls when it needs clarification or additional information before it can proceed.
The agent may call RapidAPI tools within a search loop capped at 15 messages, followed by one final synthesis call if the budget is exhausted (Appendix~\ref{sec:protocol-details}).
Tool responses come from StableToolBench's cached RapidAPI pool where available and from an LLM-simulated response otherwise.
The resulting dialogue also depends on the simulator's sampled replies and follow-up requests.
The user simulator never sees tool calls or tool responses.
Each dialogue has three requests, the initial query and two follow-ups;  the message streams and the turn and search budgets are in Appendix~\ref{sec:protocol-details}.

\subsection{User Simulation}
\label{sec:benchmark-user-sim}

The user simulator answers the agent's clarification questions from the problem's withheld context and generates the follow-up requests that continue the dialogue. It is implemented as two separate LLM-prompted generators that share the same backbone model and the same view of the conversation; the active generator is chosen by whether the agent called \texttt{finish} or \texttt{message\_user}.

\myparagraph{Follow-up query generator.}
Active after the agent calls \texttt{finish}.
The prompt instructs the simulator to extend (not rephrase) the conversation: ask a creative follow-up that builds on what the assistant already provided, explores a new facet of the problem, and is solvable by exactly one operation from \texttt{api\_list}.
A strict anti-hallucination rule forbids the simulator from attributing facts (IDs, names, numbers, dates) to the assistant or tools unless those facts appeared explicitly in the dialogue, while still permitting the user to introduce their own preferences and constraints.

\myparagraph{User-response generator.}
Active after the agent calls \texttt{message\_user}.
The prompt is the same anti-hallucination contract as above, with one important addition: the generator receives the problem's withheld context (\autoref{sec:bench-problem-design}) and is instructed to weave relevant context keywords into the reply naturally when answering the assistant's clarification, rather than dumping all keywords up front.
This is the mechanism by which the user simulator reveals the deliberately omitted details (location, dates, entity names, etc.) incrementally as the agent asks.

\myparagraph{Backbone model.}
The simulator backbone is configurable to any OpenAI-compatible model; in our main experiments (\autoref{sec:experiments}) we use Qwen3-32B-AWQ served via vLLM, an open-weight model chosen for its balance of user-simulation quality and serving cost (\autoref{sec:exp-revision}).
As a user simulator, GPT-4o-mini appears only as a closed-source comparison condition in the ablations of Appendix~\ref{sec:exp-revision} and Appendix~\ref{sec:exp-human-comparison}.

\subsubsection{Checker--Fixer Pipeline}
\label{sec:user-sim-checker-fixer}

The simulator generates each user turn (follow-up question or response to the assistant) with a single LLM call by default.
When the optional checker--fixer pipeline is enabled, two further stages run before the turn is committed.

\myparagraph{Checker.}
A separate LLM call evaluates the generated message against a fixed set of rules over the current dialogue history.
The checker detects eleven violation types (\autoref{tab:checker-violations}).
The checker returns a structured report with a boolean \texttt{has\_issues} flag and a typed list of detected violations.

\myparagraph{Fixer.}
If \texttt{has\_issues} is true, a second LLM call revises the user message.
The fixer receives the dialogue history, the original message, and the checker report, and is instructed to resolve all reported violations while preserving the communicative intent and without introducing new violations (e.g.\ no fabricating tool facts).
The revised message replaces the original and is used as the next user turn; if no issues are reported the original message is used as-is.
The same LLM is used for generation, checking, and fixing. 
Alongside the eleven LLM-judged violation types, a deterministic rule set flags near-duplicate sentence frames, verbatim repetition of prior turns, and empty or non-advancing messages; these rules require no model call and are reused as an independent measurement instrument in Appendix~\ref{sec:exp-revision-turnlevel}.
The pipeline's effect on turn quality is summarized in \autoref{sec:user-sim-validation}; the dialogue-level comparison and the paired turn-level analysis are in Appendix~\ref{sec:exp-revision}.

\subsection{Evaluation Design}
\label{sec:benchmark-evaluation}

We evaluate each generated dialogue along two axes: \myemph{dialogue quality}, measured by a five-aspect LLM-judged rubric, and \myemph{task success}, a per-subtask binary judgment of whether the user's information need for that query was fully resolved.
Both rely on LLM judges with structured function-calling outputs and a fixed prompt set across conditions.

\subsubsection{Dialogue Quality Rubric}
\label{sec:dialogue-quality-rubric}

The dialogue quality rubric scores each generated dialogue across five aspects, one LLM call per aspect:
\myemph{consistency}, factual coherence between the assistant's claims and tool outputs;
\myemph{relevance}, whether assistant turns address the user's request;
\myemph{efficiency}, economy of tool calls and turns;
\myemph{correctness}, whether the final answer is factually right with respect to tool evidence; and
\myemph{adaptability}, handling of clarifications, corrections, and underspecified inputs.
Each aspect is scored on a discrete $[0,\,1]$ scale with five levels (0, 0.25, 0.5, 0.75, 1.0) and the judge additionally returns a brief justification.
Aspect scores are combined into a single weighted aggregate (correctness 0.30; consistency, relevance, efficiency 0.20 each; adaptability 0.10). 
Because scores take only five discrete values, mean differences at the second decimal place correspond to a one-level shift on a small fraction of dialogues, so we do not read small second-decimal gaps as meaningful differences.

\subsubsection{Task Success}
\label{sec:task-success-eval}

Task success is determined by a \myemph{hybrid evaluator} that applies a deterministic hard-fail gate before invoking an LLM judge. It scores each \myemph{subtask} of the dialogue (the initial query and each follow-up query, \autoref{sec:bench-dialogue-protocol}) independently, rather than issuing a single judgment for the dialogue as a whole.
The gate handles failure modes that can be identified cheaply and with certainty, so the LLM judge is reserved for cases where the outcome is uncertain.

\myparagraph{Subtask segmentation.}
Each dialogue is split into its three main queries (one initial query and \texttt{num\_follow\_up\_queries}\,=\,2 follow-ups), and the gate and LLM judge below are applied separately to each subtask's slice of the dialogue (its query, turns, tool calls, and tool responses).

\myparagraph{Deterministic gate.}
A subtask fails without an LLM call when every substantive tool observation is StableToolBench documentation-placeholder prose, a description of the API rather than live data (for example, \textit{``The API call to \ldots\ is designed to provide \ldots''}), and none of the observations yields data-shaped output such as structured fields, lists, or typed scalars beyond bookkeeping keys like \texttt{status}, \texttt{success}, or \texttt{code}.
The gate is rarely decisive: it applies to only 20 of the 70{,}740 subtasks across the nine agents in \autoref{sec:exp-agent-backends} (0.03\%; at most five per agent), placeholder prose appears at all in only 154 (0.2\%), and no pooled subtask pass rate in \autoref{tab:agent-quality} moves by more than 0.03 points with or without it, so cache coverage does not account for the failure rates we report.

\myparagraph{LLM judge.}
If the gate passes, a single LLM call decides semantic task completion for that subtask via structured function-calling.
The judge receives the subtask's query and its slice of the dialogue (turns, tool calls and their responses, and the assistant's answer) and must return a boolean \texttt{task\_succeeded} with a brief justification.
The system prompt instructs the judge to be strict on wrong or missing answers; a subtask passes only if the user's information need for that query was \myemph{fully} met by the final outcome.

\myparagraph{Dialogue-level score.}
A dialogue's task-success score is the fraction of its subtasks that succeed, \texttt{success\_count / total\_subtasks} (e.g.\ 2/3 if the initial query and one of the two follow-ups succeed); a dialogue counts as fully successful only if every subtask succeeds.
We report both quantities in \autoref{sec:exp-agent-backends}: the pooled subtask pass rate as the headline figure, and the stricter all-subtasks-pass rate alongside it.

\section{Experiments}
\label{sec:experiments}

We report two experiments: a comparison of nine agent models on dialogue quality and task success (\autoref{sec:exp-agent-backends}), and a validation of the simulated user's turn quality (\autoref{sec:user-sim-validation}).

\subsection{Agent Comparison}
\label{sec:exp-agent-backends}

\subsubsection{Experimental Setup}
We evaluate how the choice of agent model affects dialogue quality and task success, holding every other component of the pipeline fixed so that observed variation is attributable to the agent.
The user simulator is \texttt{Qwen/Qwen3-32B-AWQ} with the checker--fixer pipeline enabled (\autoref{sec:user-sim-checker-fixer}). This configuration is validated in \autoref{sec:user-sim-validation}. The same model supplies API responses missing from the StableToolBench cache.
All rubric and task-success judge calls use \texttt{gpt-oss} at temperature 0.2 with a fixed prompt set across all agents.

\subsubsection{Agent Models Under Evaluation}
We evaluate nine agent models drawn from two access tiers.

\myparagraph{Proprietary models.}
We include four proprietary models: GPT-4o-mini, GPT-4.1-mini, GPT-5.4-mini, and Claude Opus~4.8.
The three OpenAI models are a practical cost tier for running this evaluation at scale, and they offer strong instruction-following and function-calling performance relative to their cost.
Opus~4.8 is a frontier model included as a reference for current proprietary capability.

\myparagraph{Open-Weight models.}
We include five open-weight models: Qwen3.5-397B-A17B, Qwen3.6-27B, Gemma~4-31B-IT, GPT-OSS-120B, and GLM-5.2-744B (Appendix~\ref{sec:agent-details}).
Together with the OpenAI tier this spans two orders of magnitude in parameter count (27B to 744B), both dense and MoE architectures, and four quantization schemes.
All five have reasoning enabled by default, and we did not disable it, so these agents plan with an internal thinking pass before emitting tool calls.

\subsubsection{Results}

\autoref{tab:agent-quality} reports mean scores across 2,620 dialogues per agent model, broken down by the five quality aspects, the weighted aggregate, and task success.
\autoref{tab:subtask-position} breaks task success down by the position of the main query within the dialogue.

\begin{table}[!t]
\centering
\small
\resizebox{\textwidth}{!}{
\begin{tabular}{lccccccccc}
\toprule
& \multicolumn{4}{c}{\textbf{Proprietary}} & \multicolumn{5}{c}{\textbf{Open-Weight}} \\
\cmidrule(lr){2-5}\cmidrule(lr){6-10}
\textbf{Metric} & \textbf{4o-mini} & \textbf{4.1-mini} & \textbf{5.4-mini} & \textbf{Opus~4.8} & \textbf{Qwen3.5} & \textbf{Qwen3.6} & \textbf{Gemma~4} & \textbf{GPT-OSS}$^{\ddagger}$ & \textbf{GLM-5.2} \\
& & & & & \small 397B & \small 27B & \small 31B & \small 120B & \small 744B \\
\midrule
Overall (weighted) & 0.646 & 0.740 & \textbf{0.817} & 0.809 & 0.683 & 0.709 & 0.759 & 0.693 & 0.685 \\
Task success & 57.0\% & 64.5\% & 56.3\% & \textbf{69.6\%} & 60.9\% & 61.0\% & 67.5\% & 55.2\% & 67.9\% \\
\quad all three pass & 25.1\% & 33.9\% & 26.0\% & \textbf{39.8\%} & 29.3\% & 29.0\% & 36.4\% & 24.9\% & 37.7\% \\
\midrule
Consistency & 0.839 & 0.866 & \textbf{0.924} & 0.898 & 0.837 & 0.858 & 0.899 & 0.873 & 0.836 \\
Relevance & 0.581 & 0.713 & 0.771 & \textbf{0.788} & 0.646 & 0.717 & 0.696 & 0.695 & 0.662 \\
Efficiency & 0.433 & 0.573 & \textbf{0.696} & 0.604 & 0.444 & 0.520 & 0.529 & 0.494 & 0.439 \\
Correctness & 0.722 & 0.798 & 0.861 & \textbf{0.866} & 0.766 & 0.732 & 0.862 & 0.708 & 0.752 \\
Adaptability & 0.569 & 0.693 & 0.802 & \textbf{0.908} & 0.685 & 0.707 & 0.754 & 0.680 & 0.716 \\
\bottomrule
\end{tabular}
}
\caption{Dialogue quality and task success per agent model, over a shared set of 2,620 problems. Bold denotes the best value per row.
         Open-weight models are labeled by family with total parameter count below; checkpoints and quantization are in Appendix~\ref{sec:agent-details}.
         The weighted aggregate, the pooled subtask pass rate, and the stricter all-three-pass rate are defined in \autoref{sec:benchmark-evaluation}; dialogues and subtasks for which the judge returned an error are excluded.
         $^{\ddagger}$GPT-OSS is also the judge model, so its scores are not judge-independent.}
\label{tab:agent-quality}
\end{table}

\myparagraph{ToolTurns is hard, especially for completing all three task queries.}
Pooled subtask pass rates fall in a 55--70\% band and no model reaches 70\%, so even for the best agent the judge finds the user's information need unmet on more than three subtasks in ten.
The stricter criterion is far more discriminating: requiring all three subtasks of a dialogue to succeed drops every model below 40\%, and the best model (Opus~4.8, 39.8\%) completes a full three-query dialogue in fewer than two cases out of five.
The gap between task success and passing all three highlights the difficulty of ToolTurns.
If subtask outcomes within a dialogue were independent, a 69.6\% subtask rate would imply a $0.696^3 \approx 34\%$ all-pass rate; the observed 39.8\% is higher, so successes and failures are positively correlated within a dialogue. Either some problems are simply harder, or an agent that establishes the right tool and context early tends to keep succeeding.
These two task-success rates nonetheless rank models almost identically (Spearman $\rho = 0.97$ across the nine models, differing only in two swaps of adjacent pairs), so the choice between them affects the reported difficulty of the benchmark far more than the ordering of the models.

\myparagraph{Rubric quality and task success measure different things.}
GPT-5.4-mini leads the weighted aggregate (0.817) and is best on consistency and efficiency, yet its task success (56.3\%) is second-worst of the nine, below GPT-4o-mini's 57.0\%.
GLM-5.2 is the opposite: it is second only to Opus~4.8 on task success by both definitions (67.9\% / 37.7\%) while placing seventh on the weighted aggregate (0.685).
The rubric and the task-success judge are therefore measuring different things.
Rubric and task-success rankings correlate at only $\rho = 0.23$, while the two task-success definitions correlate at $\rho = 0.97$.
The two can coincide, with Opus~4.8 best on task success and a close second on the aggregate (0.809), but the rubric does not reliably identify such models: it places GPT-5.4-mini ahead of Opus~4.8 despite a 13.3-point deficit in task success.
Consistency and efficiency both score locally well-formed, economical tool use, which is not the same as resolving the request.
An agent that issues few, coherent calls and writes a fluent answer scores well on both whether or not the answer is what the user asked for.
GPT-5.4-mini makes 3.6 API calls per dialogue, the fewest of any agent, and 6.9\% of its \texttt{finish} calls give up on the request; \autoref{sec:failure-example-rubric} shows a problem on which it scores 0.925 on the rubric and fails all three subtasks.
 The two also contribute very unequally to the aggregate.
Consistency is uniformly high and spans only 0.09 (0.836--0.924), so once a tool result is in context, maintaining factual coherence with it barely separates these models.
Efficiency spreads 0.26 (from 0.433 to 0.696) at a 0.20 weight, making it the single largest source of differences in the aggregate. It is also the aspect on which GPT-5.4-mini's lead is largest (0.092 over the runner-up, Opus~4.8).
Adaptability spreads wider still (0.34, from 0.569 to 0.908) but carries the smallest weight (0.10).

\myparagraph{Open-weight models are competitive, and parameter count predicts little.}
The two best open-weight models trail the frontier proprietary model by 1.7 and 2.1 points of task success (GLM-5.2 67.9\%, Gemma~4 67.5\%, against 69.6\% for Opus~4.8) and beat every OpenAI model (the best, GPT-4.1-mini, reaches 64.5\%); four of the five beat two of the three OpenAI models.
Gemma~4 also achieves the second-best correctness score of any model (0.862, behind only Opus~4.8 at 0.866), despite being the second-smallest open-weight model and served at 4-bit.
Across the 27B--744B range spanned by the open-weight models, size is a weak predictor of either metric.
Gemma~4 at 31B is within 0.4 points of task success of GLM-5.2 at 744B, roughly twenty-four times its size, and ahead of the 397B Qwen3.5 by 6.6 points.
The sharpest comparison is Qwen3.6-27B against Qwen3.5-397B: the newer, smaller, dense model matches the older sparse flagship on task success (61.0\% vs.\ 60.9\%) and is slightly ahead on the weighted aggregate (0.709 vs.\ 0.683), at a small fraction of the serving cost.
What separates models appears to be the quality of their dialogue and tool-selection behavior rather than raw capacity, which is what we would want from a benchmark aimed at that skill.
A locally served open-weight model is competitive on task success and on the dialogue-quality rubric.

\begin{table}[t]
\centering
\small
\begin{tabular}{lrrrr}
\toprule
\textbf{Model} & \textbf{Initial} & \textbf{FU 1} & \textbf{FU 2} & \textbf{$\Delta$} \\
\midrule
GPT-4o-mini & 65.2 & 55.7 & 50.3 & $-14.9$ \\
GPT-4.1-mini & 67.7 & 63.9 & 62.0 & $-5.7$ \\
GPT-5.4-mini & 57.6 & 55.9 & 55.5 & $-2.1$ \\
Opus~4.8 & 74.7 & 65.4 & 68.6 & $-6.1$ \\
\midrule
Qwen3.5-397B & 67.6 & 57.8 & 57.5 & $-10.1$ \\
Qwen3.6-27B & 65.8 & 57.2 & 60.1 & $-5.7$ \\
Gemma~4-31B & 73.5 & 65.5 & 63.5 & $-10.0$ \\
GPT-OSS-120B & 65.2 & 51.5 & 49.0 & $-16.2$ \\
GLM-5.2-744B & 74.1 & 64.8 & 64.8 & $-9.3$ \\
\bottomrule
\end{tabular}
\caption{Subtask success rate (\%) by position of the main query within the dialogue: the initial query and the two follow-ups (FU 1, FU 2). $\Delta$ is the change from the initial query to the second follow-up. Every model degrades as the dialogue lengthens.}
\label{tab:subtask-position}
\end{table}

\myparagraph{Follow-up success scores are lower.}
Because task success is scored per subtask, we can compare the existing evaluation protocol's scores across dialogue positions.
\autoref{tab:subtask-position} shows that all nine models score highest on the initial query and lower on both follow-ups.
The magnitude varies substantially: GPT-OSS and GPT-4o-mini lose 16.2 and 14.9 points from the initial query to the second follow-up, whereas GPT-5.4-mini loses only 2.1.
Opus~4.8 has the highest score at both ends of the dialogue (74.7\% initial, 68.6\% on the second follow-up) and loses 6.1 points, less than GLM-5.2 or Gemma~4.
For every model the larger share of the loss occurs at the \emph{first} follow-up, and the second follow-up is then roughly flat relative to it (Qwen3.6-27B and Opus~4.8 even recover 2.9 and 3.2 points).
Under this protocol, opening, context-free requests score 2--16 points higher than second follow-up requests that depend on the dialogue so far.
This pattern does not isolate a cost of conversational dependence: the judge omits prior-subtask tool evidence, which can hide the provenance of reused information (\autoref{sec:failure-example-judge}).
The effect of that omission on follow-up declines and model comparisons has not been established.
GPT-5.4-mini has consistently low scores and the \emph{flattest} position profile here, including the lowest initial-query score (57.6\%, against 73.5--74.7\% for Gemma~4, GLM-5.2, and Opus~4.8); this smaller measured decline does not establish stronger multi-turn ability.
Part of GPT-OSS's decline reflects protocol non-compliance rather than dialogue ability: in 56.1\% of its dialogues it returns an answer as plain text instead of calling \texttt{finish}, which leaves the subtask open (\autoref{sec:behaviour-stats}).

\subsection{Validating Simulated User Turns}
\label{sec:user-sim-validation}

Because the simulated user shapes the evaluation trajectory, we validate its turn quality.
On the turns it rewrites, the revision pipeline removes 41.2\% of automatically detected defects, and blind human annotators prefer the revised turn 63.3\% of the time.
These measurements use 1{,}272 user turns from a separate 322-dialogue single-call condition, rather than the 2{,}620-problem ToolTurns set.
The checker flags 543 turns (42.7\%), and the fixer rewrites 536 (42.1\% of all turns).
On those 536 rewritten turns, detected defects fall from 335 to 197; the paired within-turn change is $-0.257$ defects per turn (95\% bootstrap CI $[-0.308, -0.207]$).
In the blind human comparison, the revised turn is preferred in 76 of 120 sampled pairs ($p = 4.5 \times 10^{-3}$).
The gains concentrate on contextual unawareness, repetition, and failure to advance the task rather than on templated phrasing, which is the most frequent trigger but is indistinguishable from chance under human preference.
Dialogue-level rubric scores are insensitive to a turn-local intervention. The scores, violation-type breakdown, comparison against human-written API-Bank dialogues, and detector calibration are reported together in Appendix~\ref{sec:user-sim-appendix}.

\subsubsection{Human Validation of the Agent Evaluators}
\label{sec:agent-judge-validation}

We validate the LLM Judge used in \autoref{tab:agent-quality} through human annotation. For binary task success, human and LLM judgments achieved
substantial agreement (Cohen's $\kappa=0.68$), validating the use of an LLM judge for evaluating agents for this task.
Across the five
dialogue-quality aspects, we see moderate agreement, with a quadratic weighted Cohen's $\kappa=0.53$.
While the agreement is lesser due to the challenges of evaluating each quality dimension, we find agreement is still significant.

\section{Conclusion}
\label{sec:conclusion}

We evaluate tool-using agents on ToolTurns: multi-turn dialogue over a broad API catalogue, with clarification and follow-up requests scored along each agent's own trajectory.
Under the existing evaluation protocol, no agent is scored as completing even two dialogues in five, and every agent receives lower success scores on both follow-ups than on the initial request.
Because rubric scores rank models differently from completion ($\rho = 0.23$), progress in this setting has to be measured as completion rather than inferred from dialogue quality.

\section{Limitations}
\label{sec:limitations}

ToolTurns scores agents on trajectories that a simulated user helps determine, so measured difficulty mixes the agent with that user's choices.
Repairing individual user turns need not move dialogue-level scores; that comparison is in Appendix~\ref{sec:user-sim-appendix}.
Tool observations come from StableToolBench's cache or from an LLM simulation on a cache miss, not from live provider calls, and task success is defined against that record.
Initial queries and withheld context are generated rather than written by people, and every dialogue is scored on a fixed three-request horizon.
The dialogue-quality rubric and the task-success judge are themselves LLM judges.
The task-success judge checks answers against the tool observations of the current subtask only, so it can penalize an agent that corrects a wrong observation or reuses evidence from an earlier subtask (\autoref{sec:failure-example-judge}).
Missing provenance can make a judge's rationale incorrect without establishing that its final success label is wrong; the effect on follow-up declines and model rankings remains unquantified.

\subsection*{Reproducibility statement}

Construction of ToolTurns is specified in \autoref{sec:benchmark}: tool-set sampling and vague-query generation in \autoref{sec:bench-problem-design}, the dialogue protocol in \autoref{sec:bench-dialogue-protocol}, and both evaluators in \autoref{sec:benchmark-evaluation}.
Appendix~\ref{sec:protocol-details} gives the message streams, the turn and search budgets, and the simulator decoding temperatures; Appendix~\ref{sec:checker-violations} lists the checker's violation types and Appendix~\ref{sec:agent-details} the open-weight checkpoints and quantization schemes.
Open-weight agents and the user simulator are served with vLLM behind an OpenAI-compatible endpoint, so swapping an agent requires no change to the generation code; judge calls use a fixed prompt set at temperature 0.2 across all agents.
Appendix~\ref{sec:failure-examples} gives worked transcripts, with the tool calls and judge justifications behind our qualitative claims.
The benchmark, code, configurations, and sanitized saved scores are available at \url{https://github.com/milad621/ToolTurns} (\href{https://github.com/milad621/ToolTurns/releases/tag/v1.0.0}{release v1.0.0}).
The release documents supported table reproduction, auxiliary-result discrepancies, historical provenance limits, and third-party asset acquisition and terms.

\subsection*{AI use statement}

In this work, we used generative AI tools to generate synthetic data, to support qualitative analysis of simulated dialogues, and to interpret experimental results, and to improve the writing of the paper.
GPT-4o-mini generates the initial queries and withheld context (\autoref{sec:bench-problem-design}).
The user simulator, checker, and fixer are language models (\autoref{sec:benchmark-user-sim}).
Dialogue quality and task success are scored by LLM judges (\autoref{sec:benchmark-evaluation}), and tool observations on a cache miss are simulated by StableToolBench \citep{guo-etal-2024-stabletoolbench}.
Formulating mathematical claims, writing proofs, developing theoretical models, and translation are not applicable to this work.
We have reviewed all AI-assisted work: prompts and judge instructions were held fixed across the agents being compared, and the measurements reported in the paper were checked against those outputs.
We take responsibility for the final content of this work, including text, claims or artifacts produced with the aid of generative AI.

\subsection*{Ethics statement}

The human study is a blind preference annotation of 120 pairs of model-generated user turns (\autoref{sec:exp-human-eval}).
Each item showed the preceding dialogue and two unlabelled candidate messages, and the annotator chose the better turn under a rubric fixed in advance.
Those turns were produced by the user simulator rather than collected from people, and the paper reports only aggregate preferences.

\myparagraph{Data terms.}
Tool schemas are the RapidAPI collection released with ToolBench \citep{qin2024toolllm}.
ToolBench distributes that dataset for research and education under the Apache License 2.0.
StableToolBench \citep{guo-etal-2024-stabletoolbench}, which supplies cached responses and the fallback simulation, is released under the same license.
Neither release transfers rights in the underlying RapidAPI listings or in content returned by third-party APIs.
Live calls remain subject to RapidAPI's terms and to each provider's terms.
Our experiments use the released schemas together with StableToolBench's cache or simulation, and do not make live provider calls or claim ownership of provider content.

\bibliography{custom}
\bibliographystyle{iclr2027_conference}

\clearpage
\appendix

\section{Notes on the benchmark comparison}
\label{sec:related-work-notes}

Sizes are corpus instances, annotated dialogues, or tasks, as labeled in \autoref{tab:related-work}; catalogue scale is not a difficulty ranking.
ToolDial's 16 labels are dialogue action types, not callable tools, and its recorded-history test split has 1{,}166 dialogues.
API-Bank's 1{,}888 training dialogues and 2{,}138 training APIs are excluded.
$\tau$-bench and $\tau^{2}$-bench counts stay per domain; the generated $\tau^{2}$ telecom pool has 2{,}285 tasks, of which 114 are evaluated.
DialogTool's training corpus (16{,}142 dialogues, 20 apps, 45 APIs) is separate from this evaluation row.
The ToolBench repository samples 200 instructions from each of six subsets; those subsets are not summed, and that sampling note is in the repository rather than the paper body.
ToolBench's 126{,}486 instruction--solution pairs are the count in the current paper and repository; an earlier preprint reports 12{,}657.
DialogTool's row uses the 900-dialogue evaluation split in its Table~2.
Its prose turn count disagrees with that table, so turn counts are omitted.
ToolBench's Tool/API scope is that released catalogue: 3{,}451 tools, 16{,}464 APIs, and 49 categories.
The ToolTurns row counts the distinct tools, endpoints, and categories that appear in the 2{,}620 problems: 1{,}827, 11{,}004, and 47 (\autoref{tab:benchmark-stats}), because seed sets with fewer than three endpoints are discarded and problems are sampled from G1--G3 (\autoref{sec:bench-problem-design}).
ToolBench's evaluation split is likewise a subset of the same catalogue; that subset is not the figure in the table.
Each of our problems exposes a subset of the catalogue (26.5 endpoints on average), and exposure does not mean every endpoint is used or required; tool responses are cached or simulated with StableToolBench \citep{guo-etal-2024-stabletoolbench}.

\section{User-Simulator Evaluation}
\label{sec:user-sim-appendix}

These analyses hold the agent and tool environment fixed and evaluate the simulated user: the dialogue-level judge, the checker--fixer comparisons, the blind human study of revised turns, and a comparison against human-written API-Bank dialogues.

\subsection{User Simulator Evaluator}
\label{sec:user-sim-eval}

The dialogue quality rubric and task-success evaluator both target the \emph{agent}: they assume a reasonable user simulator and measure how well the assistant performs.
To assess the simulator independently — for example, when comparing user-simulator configurations with the agent held fixed — we use a separate LLM judge that scores complete dialogues across seven dimensions on a continuous $[0,\,1]$ scale.
The judge receives the original query, the complete dialogue, and the final assistant answer; for each dimension it returns a score and a brief justification.
The dimensions and their weights in the aggregate score are:
\myemph{realism} (0.15) — whether the simulator's behaviour and language feel like a real user;
\myemph{contextual awareness} (0.20) — whether messages are grounded in what the assistant actually said;
\myemph{relevance} (0.15) — whether follow-ups are useful and on-topic;
\myemph{natural language} (0.10) — whether phrasing is human-like and varied;
\myemph{conversational flow} (0.15) — whether the dialogue progresses naturally;
\myemph{clarity} (0.10) — whether queries are clear and interpretable; and
\myemph{goal progression} (0.15) — whether user turns help resolve the original query.
The overall score is the weighted sum of the seven dimension scores.

\subsection{Improving User Simulation with a Checker--Fixer Pipeline}
\label{sec:exp-revision}

We evaluate the checker--fixer pipeline introduced in \autoref{sec:user-sim-checker-fixer} against single-call baselines, using the dedicated user-simulator evaluator from Appendix~\ref{sec:user-sim-eval}.
This isolates the effect of the revision pipeline (and, separately, the choice of simulator LLM) on user-simulator quality with the agent and tool environment held fixed.

\subsubsection{Experimental Setup}
We compare three user-simulator conditions on a shared set of 322 dialogues, holding the agent and tool environment fixed:

\begin{itemize}
  \item \textbf{GPT-4o-mini (baseline).} The user simulator is GPT-4o-mini with a single call per turn; no checker or fixer. This serves as the closed-source baseline.
  \item \textbf{Qwen3-32B single-call.} The user simulator is Qwen3-32B-AWQ with one call per turn and no revision. This isolates the open-weight model's raw quality.
  \item \textbf{Qwen3-32B + checker--fixer.} The user simulator is Qwen3-32B-AWQ with the checker--fixer pipeline enabled.
\end{itemize}

The 322 initial queries were constructed by running the automatic vague-query procedure twice over ToolBench's three retrieval groups: 129 from G1, 129 from G2, and 64 from G3.
As in \autoref{sec:bench-problem-design}, each query is paired with all endpoints of the tools in its sampled set---one tool for G1, and possibly several for G2 and G3--- and GPT-4o-mini wrote the API-agnostic questions and their withheld context.
The agent and tool environment are identical across conditions; only the user model and the optional revision step differ.
All user-simulator evaluator calls use GPT-4o-mini as the judge with a fixed prompt set across conditions.

\subsubsection{Results}

\autoref{tab:usersim-quality} reports mean scores across 322 dialogues for each condition.

\begin{table}[!t]
\centering
\small
\begin{tabular}{lccc}
\toprule
\textbf{Dimension} & \textbf{GPT-4o-mini} & \textbf{Qwen3-32B (single-call)} & \textbf{Qwen3-32B (checker--fixer)} \\
\midrule
Overall (weighted) & 0.858 & 0.862 & \textbf{0.871} \\
\midrule
Realism & 0.807 & 0.805 & \textbf{0.812} \\
Contextual awareness & 0.863 & 0.869 & \textbf{0.880} \\
Relevance & 0.904 & 0.900 & \textbf{0.911} \\
Natural language & 0.874 & 0.877 & \textbf{0.880} \\
Conversational flow & 0.808 & 0.827 & \textbf{0.839} \\
Clarity & 0.905 & 0.902 & \textbf{0.907} \\
Goal progression & 0.864 & 0.867 & \textbf{0.877} \\
\bottomrule
\end{tabular}
\caption{User-simulator quality scores under three conditions (322 dialogues each).
         Bold denotes the best value per row.
         Qwen3 refers to Qwen3-32B-AWQ; \myemph{checker--fixer} denotes the full checker--fixer pipeline (\autoref{sec:user-sim-checker-fixer}).
         Dimensions and weights follow Appendix~\ref{sec:user-sim-eval}.}
\label{tab:usersim-quality}
\end{table}

All three conditions score in a narrow band (0.858--0.871 overall), and the differences between them are small relative to the resolution of the judge.
The checker--fixer condition is nominally highest on all seven dimensions and on the weighted overall score, and the open-weight simulator with revision (0.871) is at least at parity with the GPT-4o-mini closed-source baseline (0.858).
We deliberately do not read the $+0.009$ margin over the single-call open-weight baseline as evidence that the pipeline works: dialogue-level aggregates are structurally insensitive to this intervention, because the fixer rewrites only the minority of turns its checker flags and any effect is then averaged over the untouched turns of the dialogue.
Appendix~\ref{sec:exp-revision-turnlevel} therefore measures the intervention where it actually operates.

\myparagraph{Aggregate parity conceals a difference in specificity.}
The near-identical overall scores of GPT-4o-mini and the open-weight conditions do not imply equivalent behaviour.
Counting \myemph{concrete references} per user turn --- quoted names, identifiers, hashtags, CamelCase tokens, and proper-noun spans, extracted by a deterministic parser --- GPT-4o-mini produces 0.98 per turn against 1.98 for Qwen3-32B, at essentially the same turn length (40.2 vs.\ 42.1 words) , and 43.1\% of its turns contain no concrete reference at all against 19.3\% for Qwen3-32B.
The proprietary baseline attains comparable rubric scores in part by asking vaguer questions, which are harder for a rubric judge to penalise; a simulator that never commits to a specific entity cannot be marked down for misusing one.
This is a limitation of aspect-style rubric judging that we return to in Appendix~\ref{sec:human-eval-further}.

\subsubsection{Turn-Level Intervention Analysis}
\label{sec:exp-revision-turnlevel}

The checker--fixer is a turn-local mechanism: it inspects one candidate user message and, if flagged, rewrites that message.
Measuring it by comparing whole dialogues across two independently sampled conditions dilutes the effect twice over --- across the untouched turns of each dialogue, and across the sampling variation between runs.
We therefore evaluate the intervention directly, and pair at the level at which it acts.

\myparagraph{Protocol.}
We take all 1{,}272 user turns produced by the Qwen3-32B single-call condition.
These are unrevised drafts from the same backbone under the same prompts, so they are exactly the inputs the pipeline would have received.
Holding the preceding dialogue history fixed, we run the production checker on each turn and, wherever it fires, the production fixer, using the same model and decoding settings as the deployed system.
Only the single user message is rewritten; no dialogue is regenerated and no agent or tool call is re-executed, so the counterfactual differs from the observed draft in exactly one respect.
We then score the original draft and the rewritten message against each other under identical history.

\myparagraph{Coverage.}
The checker fires on 543 of 1{,}272 turns (42.7\%), and the fixer alters the message on 536 of those (98.7\% of firings; 42.1\% of all turns).
The most frequently reported violations in unrevised drafts are templating (75.9\% of firings, 32.4\% of all turns), lack of progress (33.7\% of firings), and contextual unawareness (27.3\%).
Templating dominates, and 39.8\% of firings report it as the sole violation, so the modal intervention is a rewording that preserves the request while breaking a reused sentence frame.
Revised turns are correspondingly close to their drafts in content but not in form: mean word-level overlap between draft and revision is 0.37.
We state this explicitly because it bounds what the pipeline should be credited with --- it is predominantly a diversity and phrasing intervention, not a hallucination filter.

\myparagraph{Result.}
On the 536 rewritten turns, we count violations using the deterministic rule set of \autoref{sec:user-sim-checker-fixer}, which is independent of any LLM judge.
Revision reduces the total from 335 to 197 detected defects, a 41.2\% reduction.
The paired within-turn change is $-0.257$ defects per turn (95\% bootstrap CI $[-0.308, -0.207]$); 167 turns improve, 37 worsen, and 332 are unchanged (two-sided sign test, $p = 6.4 \times 10^{-21}$).
Because the same detector is applied to the draft and its revision on identical history, this comparison is unaffected by the detector's absolute calibration: any constant bias cancels within the pair.

The result establishes that revision repairs the turns it targets; it does not establish a proportional improvement in end-to-end dialogue quality, which \autoref{tab:usersim-quality} shows to be small.
An intervention that fires on 42\% of turns and repairs a subset of defects in those turns should be expected to move a dialogue-level weighted average only slightly.

\subsection{User-Simulator Evaluation vs.\ Human Judgment}
\label{sec:exp-human-eval}

Appendix~\ref{sec:exp-revision-turnlevel} establishes the effect of the checker--fixer pipeline using a deterministic detector that shares its rule vocabulary with the checker itself.
That circularity is worth taking seriously, so this section evaluates the same intervention against human judgment.
Further analyses of the same labels are in Appendix~\ref{sec:human-eval-further}.

\subsubsection{Experimental Setup}

The checker--fixer acts on individual turns, so we annotate at that granularity.
From the 536 turns revised in Appendix~\ref{sec:exp-revision-turnlevel} we sampled 120 uniformly at random.
For each we presented the preceding dialogue history followed by the original draft and the revision as unlabelled candidates, in an order randomised per item.
Annotation was blind: system identities were withheld, as was any indication of which candidate had been revised.
The better user turn was selected under a rubric fixed in advance, applied in priority order: grounding in what the assistant actually provided, contextual awareness, task progression, avoidance of templated phrasing, and naturalness of user voice.
Length was explicitly excluded as a criterion, as was surface politeness unless it accurately reflected what the assistant had said or done.
No tie option was offered.
These labels calibrate the automatic measure rather than serve as ground truth.

\subsubsection{Results}

\myparagraph{Human judgment confirms the intervention.}
The revised turn was preferred in 76 of 120 cases (63.3\%, 95\% CI $[54.2, 71.7]$, two-sided sign test $p = 4.5 \times 10^{-3}$), with no position bias (side A chosen 45.0\% of the time, and side A carried the revision in 48.3\% of items).
This is an independent confirmation of Appendix~\ref{sec:exp-revision-turnlevel}: the two measurements share the sampled turns but nothing else, since the human rubric is applied to the turn as a whole rather than to the rule set the checker uses.
The recurring failures identified in unrevised drafts were verbatim reuse of a prior turn's sentence frame, requests for information the assistant had just declined or already supplied, and references to APIs or values appearing nowhere in the dialogue.

\myparagraph{The benefit is concentrated on substantive defects, not on the most frequent one.}
\autoref{tab:usersim-turnlevel-human} breaks the preference down by what the checker reported.
Where the checker caught contextual unawareness, failure to advance the task, or repetition, the revision is preferred roughly three times in four.
Where templated phrasing was the \emph{sole} violation --- 39.8\% of all firings --- the preference is indistinguishable from chance.
Verbosity triggers are worse than neutral: the original draft was preferred in 14 of 19 such cases, suggesting the fixer's response to a verbosity flag is to compress away content that human judgment valued.
Excluding verbosity triggers raises the overall preference to 70.3\% ($n=101$, $p = 5.5 \times 10^{-5}$).
The pipeline's headline firing statistic and its human-visible benefit therefore come apart, and a checker that ignored surface phrasing would plausibly retain most of the gain at lower cost.

\begin{table}[t]
\centering
\small
\begin{tabular}{lrr}
\toprule
Checker reported & Revision preferred & $p$ \\
\midrule
Contextual unawareness & 26/34 \ (76.5\%) & 0.003 \\
Repetition & 13/17 \ (76.5\%) & 0.049 \\
Lack of progress & 34/45 \ (75.6\%) & $8\!\times\!10^{-4}$ \\
Any non-templating & 49/73 \ (67.1\%) & 0.005 \\
Templating alone & 27/47 \ (57.4\%) & 0.38 \\
Verbosity & \ 5/19 \ (26.3\%) & 0.064 \\
\midrule
All revised turns & 76/120 (63.3\%) & 0.005 \\
\bottomrule
\end{tabular}
\caption{Blind human preference for the revised user turn over the original draft, by the violation type the checker reported. Categories overlap, since a turn may trigger several. Two-sided sign tests.}
\label{tab:usersim-turnlevel-human}
\end{table}

\myparagraph{Limitations.}
The human annotation covers 120 turns drawn from a single intervention, and the rubric's templating criterion overlaps with a category the checker fires on, so annotation and detector are not wholly independent on that dimension.
The turn-level design also cannot speak to whether repairing individual turns compounds into better dialogues; \autoref{tab:usersim-quality} suggests the aggregate effect is small.
We did not evaluate a checker restricted to the violation types that human annotation validates, since doing so would require regenerating the full dialogue corpus.

\subsubsection{Further analyses of the human turn comparison}
\label{sec:human-eval-further}

\myparagraph{A stylistic asymmetry that is a mechanism, not an artefact.}
Revised turns open with an acknowledgment of the assistant's previous message far more often than drafts (43.3\% vs.\ 20.0\%), and the acknowledgment-bearing candidate was chosen in 41 of the 46 items where exactly one side had one.
Read in isolation this looks like a learnable cue that could substitute for judgment.
It is better explained as the repair itself.
The 46 acknowledgment-asymmetric items more often carry contextual-unawareness flags (50.0\%, vs.\ 14.9\% among the 74 matched items) or lack-of-progress flags (67.4\% vs.\ 18.9\%).
Conversely, templating is flagged more often among the matched items (83.8\%) than among the asymmetric items (56.5\%).
These percentages count \emph{any} templating flag, including flags that co-occur with other violations: 88 of the 120 items have a templating flag, but only 47 have templating as their sole flag (\autoref{tab:usersim-turnlevel-human}).
Opening a turn by registering what the assistant just said is one way to repair a turn that ignored it, so conditioning on acknowledgment may also remove part of the intervention's effect.
The acknowledgment-matched and templating-only subsets therefore should not be treated as interchangeable.
We report the asymmetry for transparency but do not treat the matched-subset rate (56.8\%, $p = 0.30$) as the corrected estimate.

\myparagraph{The deterministic detector is directionally sound but under-sensitive.}
Human labels let us calibrate the instrument used in Appendix~\ref{sec:exp-revision-turnlevel}.
Across the 120 annotated pairs the deterministic count expresses a preference on only 47; on those it agrees with the human choice 68.1\% of the time ($\kappa = 0.23$) and favours the revision 36 to 11, matching the human direction.
On the remaining 73 it is silent, yet the revision is preferred on 45 of them (61.6\%, $p = 0.06$).
The detector thus misses roughly three fifths of the changes a human perceives while rarely pointing the wrong way.

\myparagraph{Implications for how these measures should be used.}
Two conclusions follow, and together they justify the design of Appendix~\ref{sec:exp-revision-turnlevel}.
First, an absolute defect rate from this detector is not a trustworthy standalone quality measure, and we do not report one; it is silent on the majority of changes a human notices.
Second, weak absolute calibration does not preclude \emph{paired} use: applied to a draft and its revision under identical history, a constant bias cancels within the pair, leaving the difference interpretable even though the levels are not.
Since the detector is silent far more often than it errs, that difference understates rather than inflates the effect --- which is why the automatic estimate ($41.2\%$ fewer defects) and the human estimate ($63.3\%$ preference) can both be reported without conflict.
More broadly, the specificity gap in Appendix~\ref{sec:exp-revision} --- GPT-4o-mini making half as many concrete references per turn while scoring comparably on the rubric --- indicates that aspect-style judges reward safe, underspecified user behaviour, and should be complemented by measures sensitive to what a simulator actually commits to.

\subsection{Comparing the User Simulator to Human-Written Dialogues (API-Bank)}
\label{sec:exp-human-comparison}

The experiments in Appendix~\ref{sec:exp-revision} compared user-simulator configurations against each other; the more fundamental question is whether the simulator behaves at all like a real user.
We use API-Bank \citep{li-etal-2023-api} as a source of human-written, multi-turn user--assistant dialogues with explicit tool-use structure, and apply the same user-simulator evaluator (Appendix~\ref{sec:user-sim-eval}) to both human-written and simulator-generated dialogues over a matched task distribution.

\subsubsection{Experimental Setup}

\myparagraph{Why API-Bank.}
API-Bank's evaluation split provides annotated conversations that --- unlike most ToolBench-style corpora, which centre on RapidAPI tool definitions and synthetic or retrieved queries --- are closer to what we want for studying tool-using dialogue under a fixed task distribution.
It is not, however, RapidAPI-isomorphic: API-Bank names capabilities with short symbolic API names (e.g.\ calendar or email operations), so it cannot be run directly against our StableToolBench environment without a bridging step to real RapidAPI tool schemas.

\myparagraph{Data preparation.}
We normalize API-Bank test data into a single repository layout: 622 JSON files (one per dialogue) plus a global \texttt{initial\_queries.json} index (622 rows; 522 multi-turn level-1/2 scenes, 50 level-3 holistic scenes, 50 level-3 batch scenes).
Each dialogue record carries \texttt{dialogue\_messages} (user / assistant roles), an \texttt{initial\_query}, \texttt{metadata.tools} (distinct tool names in first-seen order), an optional \texttt{api\_request\_trace}, and gold fields where the source provides them (e.g.\ next API string vs.\ next utterance, depending on level).
This format is stable for downstream scripting and reusable for the LLM-judge comparison below.

\myparagraph{Mapping API-Bank tasks to RapidAPI tools.}
Because our agent environment executes RapidAPI-style tools indexed from a local ToolBench tree (category / tool / API JSON), we map each API-Bank dialogue's tool hints into a small set of concrete RapidAPI operations.
We first collect the distinct API-Bank tool names appearing in \texttt{metadata.tools} across the slice we use (68 unique names in the full 622-dialogue set).
For each name, we retrieve candidates from the local catalog via embedding similarity over serialized tool+API descriptions, then optionally apply a single lightweight LLM rerank that chooses a fixed number of candidates by index, so the model cannot invent tool names.
For each dialogue, we merge per-name lists in hint order, deduplicate by \texttt{(category, tool, API)}, and cap the number of exposed APIs (top 15), producing a merged query file aligned with our existing dialogue runs (\texttt{original\_query}, \texttt{query\_str}, \texttt{context}, \texttt{api\_list}).

\myparagraph{Simulated dialogues.}
For each of the 622 API-Bank initial user goals we run our generative dialogue protocol: a tool-using agent and a user simulator alternate until dialogue limits or termination criteria are met.
The agent may call RapidAPI tools through StableToolBench, restricted to the mapped \texttt{api\_list} for that instance.
We generate two simulator conditions --- GPT-4o-mini and Qwen3-32B-AWQ with the checker--fixer pipeline enabled --- holding the agent and tool environment fixed.

\myparagraph{Evaluation procedure.}
We score the user side of each dialogue with the user-simulator evaluator from Appendix~\ref{sec:user-sim-eval}: same judge model (GPT-4o-mini), same prompt, same seven dimensions, applied uniformly to all four dialogue sets.
The two human reference slices are \myemph{Human (all 622)} and \myemph{Human (excl.\ 54)}, where the latter removes 54 purely one-line items that do not exercise multi-turn behaviour.
The aim is to determine whether, under this single rubric, the simulator is not much worse than real users --- justifying its use for larger-scale experiments --- \emph{not} to claim the judge is ground truth or that the two sets differ only in who wrote the user turns.
In particular, the human references are tied to API-Bank's assistant and environment, while our simulated runs use our agent and the mapped RapidAPI tools, so part of any gap may reflect the assistant and environment rather than the user model alone.

\subsubsection{Results}

\autoref{tab:human-comparison} reports user-simulator quality scores for the four dialogue sets, each judged by the same LLM under identical instructions.

\begin{table}[!t]
\centering
\small
\setlength{\tabcolsep}{5.5pt}
\begin{tabular}{lcccc}
\toprule
\textbf{Dimension} & \textbf{GPT-4o-mini user} & \textbf{Qwen3-32B user} & \textbf{Human (all 622)} & \textbf{Human (excl.\ 54)} \\
\midrule
Overall (weighted) & 0.845 & \textbf{0.851} & 0.737 & 0.772 \\
\midrule
Realism & 0.762 & \textbf{0.766} & 0.576 & 0.610 \\
Contextual awareness & 0.900 & 0.888 & \textbf{0.915} & 0.909 \\
Relevance & 0.864 & \textbf{0.885} & 0.618 & 0.677 \\
Natural language & 0.861 & \textbf{0.867} & 0.776 & 0.823 \\
Conversational flow & 0.783 & \textbf{0.807} & 0.710 & 0.772 \\
Clarity & 0.915 & 0.903 & 0.914 & \textbf{0.916} \\
Goal progression & 0.841 & \textbf{0.850} & 0.663 & 0.715 \\
\bottomrule
\end{tabular}
\caption{User-simulator quality scores for our simulator vs.\ human-written API-Bank dialogues. Each simulator condition and the full human reference contain 622 records; the filtered human reference contains 568. All are scored by the same LLM judge with the same prompt and dimensions (Appendix~\ref{sec:user-sim-eval}).
         \myemph{Qwen3-32B user} uses the checker--fixer pipeline.
         \myemph{Human (excl.\ 54)} removes 54 purely one-line items from the reference set.
         Bold denotes the best value per row.
         Simulator and human dialogues differ in the assistant and tool environment as well as in the user, so observed gaps reflect protocol differences and not user behaviour alone.}
\label{tab:human-comparison}
\end{table}

Simulated users score higher on average than the filtered reference set (0.851 vs.\ 0.772 overall).
Part of this gap is depth: our generated dialogues typically include many user turns and follow-ups, whereas the reference bundle is mixed --- shorter threads and single-turn-style items remain common even after the 54 purely one-line cases are removed.
The rubric rewards coherent follow-up behaviour and sustained progress, so longer, follow-up-heavy runs are favoured even when both sides are judged by the same template.
The remaining contrast reflects style and packaging: reference text often reads like benchmark snippets (economical wording, occasionally unnatural disclosure or formatting), while simulated users tend to be more cooperative and verbose in ways the automated judge reads as fluent and on-task.
The gap should therefore be read as distributional and protocol mismatch --- especially turn count and elaboration --- not as proof that synthetic users are uniformly ``more human'' under a single gold standard.
Notably, on the two dimensions least sensitive to dialogue length --- \emph{contextual awareness} and \emph{clarity} --- humans and the simulators score within 0.01--0.03 of each other across all four sets, suggesting parity on surface-level grounding once length effects are factored out.

\section{Dialogue generation details}
\label{sec:protocol-details}

\myparagraph{Dual message streams.}
The runtime maintains two parallel message lists:
\myemph{\texttt{search\_messages}} is the agent's full view, including system prompt, tool calls, tool responses, and intermediate reasoning, and is carried across turns so the agent has full context;
\myemph{\texttt{dialogue\_messages}} is the user-visible conversation, containing only the initial query, user-simulator turns, and the agent's terminating \texttt{finish}/\texttt{message\_user} content.
The user simulator sees only \texttt{dialogue\_messages} --- never tool calls or tool responses --- which is what enforces realism: the simulator behaves like a human user who does not observe the assistant's internal tool operations.

\myparagraph{Termination and message budgets.}
Each dialogue consists of exactly one initial query followed by a fixed \texttt{num\_follow\_up\_queries}\,=\,2 follow-up queries, giving three main queries (\myemph{subtasks}) per dialogue.
Each main query's sub-dialogue is capped at \texttt{max\_turns\_per\_main\_query}\,=\,10 user-visible messages; if the cap is reached before the agent calls \texttt{finish} or \texttt{message\_user}, the agent is forced to conclude that subtask.
The response generator may still emit the sentinel \texttt{<STOP>}, treated as the user abandoning the current subtask early.
A separate budget, \texttt{max\_search\_messages}\,=\,15, caps the agent's within-turn search depth; if exceeded, the agent is given one synthesis call that prompts it to call \texttt{finish} based on the tool responses collected so far, rather than aborting the turn outright.

\myparagraph{Follow-up generator output.}
The follow-up generator returns a JSON object with the new \texttt{instruction} text plus a candidate \texttt{api\_calls} hint (used only for inspection; the agent re-plans freely).

\myparagraph{Decoding temperatures.}
The simulator generates each turn at temperature 0.7 for follow-up queries and 0.8 for clarification responses (slightly higher for variety).

\section{Checker violation types}
\label{sec:checker-violations}

\begin{table}[H]
\centering
\small
\begin{tabularx}{\textwidth}{lX}
\toprule
\textbf{Type} & \textbf{What the checker flags} \\
\midrule
Repetition & Repeating prior requests or information already given \\
Hallucination & Fabricating facts attributed to the assistant or tools \\
Contextual unawareness & Ignoring what the assistant actually said or asking for already-provided information \\
Verbosity & Unnecessary or overly long text \\
Conversational flow & Unnatural or disconnected phrasing \\
False agreement & Claiming to have received information that was not provided \\
Excessive restatement & Echoing the assistant's reply instead of a brief acknowledgment \\
Lack of progress & Not answering the assistant's question or not advancing the task \\
Vagueness & Queries too generic to be actionable \\
Ignoring assistant questions & Not addressing a direct question \\
Templating & Reusing a near-verbatim sentence frame from an earlier user turn with only the key term swapped \\
\bottomrule
\end{tabularx}
\caption{The eleven violation types detected by the user-turn checker (\autoref{sec:user-sim-checker-fixer}).}
\label{tab:checker-violations}
\end{table}

\section{Open-weight agent checkpoints}
\label{sec:agent-details}

The five open-weight agents in \autoref{sec:exp-agent-backends} are
\emph{Qwen3.5-397B-A17B} (sparse MoE, 397B total / 17B active, official FP8),
\emph{Qwen3.6-27B} (dense, bf16),
\emph{Gemma~4-31B-IT} (dense, 4-bit QAT \texttt{w4a16}),
\emph{GPT-OSS-120B} (sparse MoE, native MXFP4), and
\emph{GLM-5.2-744B} (NVFP4).
Routing all models through the same OpenAI-compatible API means no changes to the dialogue generation code are required when swapping agents.

\section{Worked examples}
\label{sec:failure-examples}

The transcripts below are taken from the nine-agent runs of \autoref{sec:exp-agent-backends}; each is identified by its problem index in ToolTurns.
Tool outputs are abridged, long agent messages are cut where marked [\ldots], and tool names are shortened by dropping the \texttt{\_for\_<tool>} suffix.
User turns, agent messages, and judge justifications are otherwise verbatim.
All tool observations come from StableToolBench's cache or its LLM simulator (\autoref{sec:bench-dialogue-protocol}), not from live APIs, and several of the examples turn on that fact.
\textsc{judge} lines give the task-success verdict for the subtask (\autoref{sec:task-success-eval}).

\subsection{One problem, several trajectories}
\label{sec:failure-example-running}

Problem 2610 is a real instance of the sketch in \autoref{sec:bench-problem-design}.
The initial query is \textit{``I heard there's a big game coming up soon. Can you tell me when it is?''}, the withheld context is \texttt{NBA}, \texttt{Los Angeles}, \texttt{Lakers}, \texttt{October}, and the tool set exposes 48 endpoints from three tools (Free NBA, Web Search, GeoDB Cities).
\autoref{tab:example-2610} gives every agent's outcome.
The NBA tool is unreliable in this environment: depending on the call it returns an error, games from past seasons, or another team's games under the Lakers' ID.
Agents therefore have to clarify the request and then decide what to do with bad tool data, and they differ on both.

\begin{table}[ht]
\centering
\small
\setlength{\tabcolsep}{1.5pt}
\begin{tabular}{lccccccccc}
\toprule
& \textbf{4o-mini} & \textbf{4.1-mini} & \textbf{5.4-mini} & \textbf{Opus~4.8} & \textbf{Qwen3.5} & \textbf{Qwen3.6} & \textbf{Gemma~4} & \textbf{GPT-OSS} & \textbf{GLM-5.2} \\
\midrule
Initial & \xmark & \xmark & \xmark & \xmark & \xmark & \cmark & \cmark & \xmark & \xmark \\
FU 1 & \xmark & \cmark & \cmark & \cmark & \xmark & \cmark & \cmark & \xmark & \cmark \\
FU 2 & \cmark & \cmark & \cmark & \xmark & \cmark & \cmark & \cmark & \xmark & \cmark \\
\midrule
Rubric overall & 0.55 & 0.75 & 0.68 & 0.80 & 0.53 & 0.85 & 0.80 & 0.43 & 0.70 \\
\bottomrule
\end{tabular}
\caption{Subtask outcomes and weighted rubric score for problem 2610. Follow-up queries differ across agents because each is generated from that agent's own dialogue.}
\label{tab:example-2610}
\end{table}

\myparagraph{Gemma~4: clarifies, then answers only from tool evidence (3/3).}
\begin{transcript}
\tuser ``I heard there's a big game coming up soon. Can you tell me when it is?''
\tcall{web\_search(q="big game coming up soon sports")}
\tagent Could you please let me know which sport or specific game you're referring to? There are many big games coming up across different leagues like the NBA, NFL, NHL, or college sports, and I'd love to give you the exact details!
\tuser Oh, sorry! I was thinking about the NBA season since I'm in Los Angeles and a huge Lakers fan. I was wondering if there's a big game coming up for them in October?
\tcall{get\_all\_teams(page="1")}
\tcall{get\_all\_games(seasons="2024", team\_ids="14")}
\ttool{Games for the Milwaukee Bucks. [\ldots]}
\tcall{web\_search(q="Los Angeles Lakers schedule October 2024 big games")}
\ttool{``The Los Angeles Lakers will open the 2024-25 season with a high-stakes road game against the Golden State Warriors on October 25, 2024. They will also host the Phoenix Suns on October 29, 2024 [\ldots] Their most anticipated game is against the Boston Celtics on October 31, 2024, at Crypto.com Arena.''}
\tagent The most anticipated matchup is on \textbf{October 31, 2024}, when the Lakers will host the \textbf{Boston Celtics} at Crypto.com Arena. Other notable games in October include: October 25, 2024: A high-stakes road game against the Golden State Warriors to open the season. October 29, 2024: A key home game against the Phoenix Suns. [\ldots]
\tjudge{\cmark\ The assistant provided specific dates for the Lakers' upcoming big games in October 2024, matching the information from websearch results and fully answering the clarified user request.}
\end{transcript}
Gemma disregards the schedule endpoint's wrong-team output, and every date in its answer appears in a tool observation.
The judge checks consistency with those observations, not with the real 2024--25 schedule.

\myparagraph{GPT-4o-mini: skips clarification and answers a different question (1/3).}
\begin{transcript}
\tuser ``I heard there's a big game coming up soon. Can you tell me when it is?''
\tcall{get\_all\_games(seasons=["2023"], per\_page="5", page="1")}
\ttool{\texttt{\{"error": "Failed to generate fake response"\}} \quad(twice)}
\tcall{news\_search(q="upcoming NBA games")}
\ttool{``The Los Angeles Lakers and Golden State Warriors are set to face off in the final round of the NBA playoffs, with the first game scheduled for July 10, 2026.''}
\tagent The first game in the NBA playoffs between the Los Angeles Lakers and the Golden State Warriors is scheduled for July 10, 2026. You can find more details in the news articles regarding upcoming NBA games.
\tuser Actually, that's not quite right - the NBA regular season starts in October, not the playoffs. The Lakers have a few key games coming up in October to kick off the season. Do you know when their first home game will be in Los Angeles this year?
\tomit{eleven further NBA-API calls: the team list, then ten requests for 2023-season games, each of which returns an error or another team's games}
\tagent I currently cannot retrieve the specific date for the Los Angeles Lakers' first home game in October. You might want to check the official NBA website or sports news sites for the latest updates on their schedule.
\tuser Oh, right! I forgot you couldn't access the schedule directly. I actually just checked the Lakers' website and their first home game in Los Angeles this season is on October 28th against the Sacramento Kings. [\ldots]
\tomit{the agent returns search-result links through \texttt{message\_user} until the subtask reaches the 10-message cap}
\tjudge{\xmark\ The assistant initially gave an incorrect date and never provided the correct date of the upcoming Lakers game; the final answer only offered streaming and ticket links without confirming the game's date, so the user's original request for when the big game is was not satisfied.}
\end{transcript}
The agent never asks which game the user means; it guesses a league, finds an unrelated event in a news snippet, and presents it as the answer.
The user's correction supplies the withheld context, but the agent gets no further with it.
The transcript also shows a simulator failure: once the agent gives up, the simulated user supplies a date and opponent that appear nowhere in the dialogue.
The anti-hallucination rule in \autoref{sec:benchmark-user-sim} forbids attributing facts to the assistant, but it does not stop the user from asserting facts of its own, and the checker did not flag this turn.
The judge still fails the subtask, since the agent itself never answered, but the remaining subtasks build on the user's invented game.

\myparagraph{GPT-OSS: clarifies correctly, then fabricates the schedule (0/3).}
\begin{transcript}
\tuser ``I heard there's a big game coming up soon. Can you tell me when it is?''
\tagent Sure! Could you let me know which sport or specific game you're interested in (e.g., NBA, football, a particular team, etc.)?
\tuser Oh, I'm talking about the NBA! Specifically, the Los Angeles Lakers. I know they have some home games in October, but I'm not exactly sure which dates. Do you know when their next game is this month?
\tomit{six NBA-API calls, none of which returns a 2026 schedule (they return errors, the team list, or games from the 2018 season), and eight web searches whose snippets say the Lakers play the Warriors, Suns, and Jazz in October but give no dates}
\tagent According to the official 2026-27 Los Angeles Lakers schedule posted on NBA.com, the Lakers' first home game in October 2026 is on \textbf{Saturday, October 25, 2026} when they host the Golden State Warriors at Crypto.com Arena. The next two home games that month are: \textbf{Monday, October 27, 2026} vs.\ the Phoenix Suns; \textbf{Thursday, October 30, 2026} vs.\ the Utah Jazz. These three games are the Lakers' only home appearances in October 2026. [\ldots]
\tjudge{\xmark\ The assistant fabricated specific October 2026 dates for Lakers home games without any supporting data from the tools; the available tool results did not contain the schedule, so the answer is unreliable and does not correctly answer the user's request for the next game.}
\tuser \textit{(follow-up 1)} Can you provide more details about the October 25th game against the Golden State Warriors, such as the start time or any special events associated with it?
\tomit{searches return three different start times; the agent reports one of them, but as plain text rather than through \texttt{finish}, so the subtask stays open for three more user turns about Fan Fest perks, merchandise, and food}
\tuser \textit{(follow-up 2)} Can you check if there are any vegetarian or gluten-free options available at the Fan Fest food court, since I have dietary restrictions and want to make sure I can enjoy the snacks without issues?
\end{transcript}
None of the three dates appears in any tool observation; the opponents do, and the attribution to NBA.com is invented.
Because the simulator sees only the user-visible conversation, it treats the fabricated schedule as fact, and both follow-ups are about a game whose date the agent made up.
The first follow-up's verdict is also affected by a protocol failure specific to this agent.
The runtime treats a turn that ends without \texttt{finish} as a message to the user, so the simulator keeps replying, and the judge scores the last answer in the slice, which is about food, against a question about tip-off time.
\autoref{tab:behaviour-stats} shows how often this happens.

\myparagraph{Opus~4.8: clarifies, answers the date, cannot confirm the time (1/3).}
Opus asks for the sport, team, and date range, finds a season opener against the Warriors on 5~October~2026 in a news snippet, and says that it could not confirm the tip-off time.
The judge fails the initial subtask because the user's clarification asked for ``the exact date and time.''
Under the strict ``fully met'' criterion, an honest partial answer receives the same score as a wrong one.

\subsection{High rubric score, no task success}
\label{sec:failure-example-rubric}

Problem 140 isolates the disagreement between the dialogue-quality rubric and task success reported in \autoref{sec:exp-agent-backends}.
The query is \textit{``I'm looking for a new smartphone but I have a specific brand in mind.''} with withheld context \texttt{Samsung}, \texttt{Galaxy S23}, \texttt{features}, and three endpoints (\texttt{GetBrands}, \texttt{GetPhonesByBrandId}, \texttt{GetPhoneDetails}).
The brand list the environment returns is truncated alphabetically at Lenovo, so Samsung's identifier never appears in any observation and no agent can complete the intended tool path.
GPT-5.4-mini receives the highest rubric score of the nine agents on this problem (0.925) and fails every subtask; Opus~4.8 scores 0.60 on the rubric and passes every subtask (\autoref{tab:example-140}).

\begin{table}[ht]
\centering
\small
\setlength{\tabcolsep}{3pt}
\begin{tabular}{lccccccc}
\toprule
& Consistency & Relevance & Efficiency & Correctness & Adaptability & Overall & Task success \\
\midrule
GPT-5.4-mini & 1.00 & 1.00 & 0.75 & 1.00 & 0.75 & 0.925 & 0/3 \\
Opus~4.8 & 1.00 & 0.25 & 0.50 & 0.50 & 1.00 & 0.600 & 3/3 \\
\bottomrule
\end{tabular}
\caption{Rubric aspects and task success for two agents on problem 140.}
\label{tab:example-140}
\end{table}

\myparagraph{GPT-5.4-mini.}
\begin{transcript}
\tagent Which brand are you interested in?
\tuser I'm looking at Samsung phones, specifically the Galaxy S23. I heard it has some great features, but I'd like to know more about it. [\ldots]
\tcall{getbrands()}
\ttool{\texttt{[\{"id":237,"name":"Acer"\ldots\},\ldots,\{"id":288,"name":"Lenovo",\ldots} \quad(truncated)}
\tagent I can look up the Galaxy S23 details, but I need the Samsung brand entry first. Would you like me to continue by finding Samsung in the brand list and then pulling the S23 specs?
\tuser Yes, please go ahead and find Samsung in the brand list. [\ldots]
\tcall{getbrands()}
\ttool{(the same truncated list)}
\tcall{finish(return\_type="give\_up\_and\_restart", final\_answer="I can look up Samsung and the Galaxy S23, but I only received a partial brand list response and couldn't reliably confirm Samsung's entry or retrieve the S23 details from it. Please try again, and I'll continue from the brand list to find Samsung and then the Galaxy S23 specs.")}
\tjudge{\xmark\ The assistant failed to provide any details about the Samsung Galaxy S23 and instead gave up, indicating it could not retrieve the necessary information.}
\tomit{both follow-ups repeat the pattern: one more \texttt{getbrands} call and another \texttt{give\_up\_and\_restart}}
\end{transcript}

\myparagraph{Opus~4.8.}
\begin{transcript}
\tomit{asks for the brand; the user names Samsung and the Galaxy S23; \texttt{getbrands} returns the same truncated list, and the agent tries brand IDs that do not appear in it}
\tcall{getphonesbybrandid(brandid=323)}
\ttool{\texttt{\{"error": "Failed to generate fake response"\}}}
\tcall{getphonesbybrandid(brandid=322)}
\ttool{\texttt{[12345, 12346, \ldots, 12400]}}
\tcall{getphonedetails(phoneid=12345)}\quad\texttt{\ldots(12346)}\quad\texttt{\ldots(12347)}
\ttool{``Phone ID 12345 corresponds to a hypothetical device. [\ldots]''; ``Phone ID 12346 corresponds to the Apple iPhone 14 Pro. [\ldots]''; the same for 12347}
\tagent [\ldots] I want to be upfront: our phone database wasn't returning reliable Samsung records just now (the entries came back either incomplete or mislabeled), so rather than give you inaccurate specs, here's a trustworthy rundown of the Galaxy S23 family: \textbf{Galaxy S23 (base model)} -- Released Feb 2023; 6.1'' Dynamic AMOLED 2X display, 120Hz [\ldots]
\tjudge{\cmark\ The assistant provided a comprehensive overview of Samsung Galaxy S23 models, covering specifications, variants, and guidance, directly satisfying the user's request for feature details despite missing database records.}
\end{transcript}
The rubric credits GPT-5.4-mini for never saying anything the tools contradict, and it marks Opus down on relevance and correctness for answering beyond them.
The task-success judge asks only whether the user's need was met, and it accepts an answer drawn from the model's own knowledge when the agent reports that the tool is broken.
We report this as a property of the two metrics, not a verdict on which agent behaved correctly.
The pattern is not specific to this problem.
In 63 dialogues GPT-5.4-mini scores at least 0.9 on the rubric and fails all three subtasks, and 6.9\% of its \texttt{finish} calls are \texttt{give\_up\_and\_restart}, against 0.1\% for Opus~4.8 (\autoref{tab:behaviour-stats}).

\subsection{Limitations of subtask-level judging}
\label{sec:failure-example-judge}

Task success is judged against the tool observations in each subtask's slice of the dialogue (\autoref{sec:task-success-eval}).
The cases below illustrate limitations of the evidence available to the judge; errors in a judge's rationale must be distinguished from verified errors in its success label.

\myparagraph{A wrong tool observation (problem 589, Opus~4.8).}
The user has chosen a meeting slot on Wednesday, 30~September~2026, 11:00--12:00 Central Time, and asks how it looks in Eastern and Pacific time.
\begin{transcript}
\tcall{time\_zone\_converter(datetime="2026-09-30T11:00:00", from\_tzname="US/Central", to\_tzname="US/Pacific")}
\ttool{\texttt{"from": \{"datetime": "2026-09-30T11:00:00-05:00"\}, "to": \{"datetime": "2026-09-30T08:00:00-08:00"\}}}
\tagent [\ldots] (all on daylight saving time that date) [\ldots] Pacific (PT / PDT): 9:00 AM -- 10:00 AM [\ldots]
\tjudge{\xmark\ The assistant gave the correct Eastern Time conversion but incorrectly listed Pacific Time as 9:00-10:00 AM instead of the correct 8:00-9:00 AM per the tool results, so the user's request was not fully met.}
\end{transcript}
The tool applies the Pacific standard-time offset in late September, when daylight saving time is in effect.
The agent overrides it with the correct answer, and the judge, which is instructed to check the answer against tool results, fails the subtask.
The rubric scores this dialogue 0.95.

\myparagraph{Evidence outside the judged slice (problem 1515, Gemma~4).}
In the initial subtask, a balance endpoint reports that the user's wallet holds ``2.50 ETH ($\approx$ \$5,000.00 USD)'' and ``3,000 USDC ($\approx$ \$3,000.00 USD)'', and the agent reports this correctly (\cmark).
\begin{transcript}
\tuser \textit{(follow-up 1)} That's helpful! Could you also show me a breakdown of the specific assets in my Ethereum wallet?
\tcall{wallets\_assets\_by\_chain(address="0xabc123def456", chain="ethereum")}
\ttool{\texttt{\{"data": [], "projects": [], "total": 0\}}}
\tagent Based on your wallet (0xabc123def456) on the Ethereum blockchain, here is the breakdown of your assets: 2.50 ETH (approximately \$5,000.00 USD); 3,000 USDC (approximately \$3,000.00 USD)
\tjudge{\xmark\ The assistant fabricated asset balances (2.50 ETH, 3,000 USDC) even though the tool returned an empty asset list, resulting in a critical hallucination and failing to provide the correct breakdown.}
\end{transcript}
The figures come from a tool observation one subtask earlier, which the judge never sees because it receives only the current subtask's slice.
Calling these figures fabricated is incorrect, but restoring their provenance does not settle whether the final answer is correct: the agent does not resolve the conflict between the two endpoints.
The reported scores retain the original success labels.
These examples do not quantify the effect of omitted prior evidence on follow-up declines or model rankings.

\subsection{Behavioural statistics}
\label{sec:behaviour-stats}

\autoref{tab:behaviour-stats} counts, over all 2{,}620 dialogues per agent, the behaviours that recur in the examples above.

\begin{table}[ht]
\centering
\small
\resizebox{\textwidth}{!}{
\begin{tabular}{lccccccccc}
\toprule
& \textbf{4o-mini} & \textbf{4.1-mini} & \textbf{5.4-mini} & \textbf{Opus~4.8} & \textbf{Qwen3.5} & \textbf{Qwen3.6} & \textbf{Gemma~4} & \textbf{GPT-OSS} & \textbf{GLM-5.2} \\
\midrule
Task success (\%) & 57.0 & 64.5 & 56.3 & 69.6 & 60.9 & 61.0 & 67.5 & 55.2 & 67.9 \\
\midrule
\texttt{finish} calls that give up (\%) & 7.1 & 1.6 & 6.9 & 0.1 & 7.9 & 6.1 & 1.1 & 12.4 & 0.6 \\
Initial queries with no API call (\%) & 1.7 & 4.0 & 28.3 & 3.3 & 2.3 & 2.2 & 4.1 & 9.2 & 1.1 \\
API calls per dialogue & 15.6 & 8.3 & 3.6 & 7.2 & 14.0 & 10.3 & 10.6 & 10.7 & 17.4 \\
\texttt{message\_user} calls per dialogue & 3.05 & 1.99 & 0.73 & 0.86 & 1.38 & 1.78 & 0.92 & 1.36 & 1.29 \\
Dialogues with an answer outside \texttt{finish} (\%) & 0.3 & 0.0 & 0.5 & $<$0.1 & $<$0.1 & 0.3 & $<$0.1 & 56.1 & $<$0.1 \\
\bottomrule
\end{tabular}
}
\caption{Agent behaviour over the 2{,}620 ToolTurns dialogues.
\myemph{Give up}: \texttt{finish} called with \texttt{return\_type} \texttt{give\_up\_and\_restart}.
\myemph{No API call}: the initial subtask, including any clarification turns, contains no call other than \texttt{finish} or \texttt{message\_user}.
\myemph{Answer outside \texttt{finish}}: at least one assistant turn that contains a \texttt{final\_answer} payload as plain text instead of a \texttt{finish} call; the runtime treats such a turn as a message to the user.}
\label{tab:behaviour-stats}
\end{table}

Across the nine agents, the give-up rate ranks them almost exactly in reverse order of task success (Spearman $\rho = -0.93$).
The three agents with the highest task success give up in 1.1\% of \texttt{finish} calls or fewer, and every other agent gives up in at least 1.6\%.
GPT-5.4-mini makes the fewest API calls (3.6 per dialogue) and handles 28.3\% of initial queries without any API call, compared with 1.1--9.2\% for the other agents; it succeeds on only 44.3\% of those.
This combination explains how it leads on efficiency while trailing on task success (\autoref{tab:agent-quality}).
GPT-OSS ends a turn with a plain-text answer instead of \texttt{finish} in 56.1\% of its dialogues, so its subtasks often stay open and are scored on whatever answer ends the slice.
Part of GPT-OSS's low task success, and of its 16.2-point follow-up decline, therefore reflects non-compliance with the tool-calling protocol rather than failures of reasoning or tool use.
GPT-4o-mini calls \texttt{message\_user} most often (3.05 per dialogue); as in \autoref{sec:failure-example-running}, some of these calls carry answers rather than questions, which also keeps a subtask open.
Finally, keyword matching on the judge's justifications (e.g.\ ``fabricated'', ``hallucination'', ``not supported by'') finds allegations of unsupported content in 7.3\% of failed initial subtasks and 16.5\% of failed follow-ups.
These heuristic rates describe judge rationales, not verified unsupported answers; omitted prior evidence can produce an incorrect rationale (\autoref{sec:failure-example-judge}).

\clearpage
\section{Prompts}
\label{sec:prompts}

This appendix records the prompts behind the reported dialogues and the agent evaluation in \autoref{sec:experiments}.
Two generators produce the simulated user turns (\autoref{sec:benchmark-user-sim}).
The agent is scored by the five-aspect dialogue-quality rubric (\autoref{sec:dialogue-quality-rubric}) and the per-subtask task-success judge (\autoref{sec:task-success-eval}).
Those six judge templates are the only prompts that evaluator loads: the code has no alternate version of this judge, and the tables in \autoref{sec:exp-agent-backends} use them with \texttt{gpt-oss} at temperature 0.2.
The separate judge that scores user-simulator quality is not included here.

Brace-wrapped names such as \texttt{\{conversation\_history\}} are filled per turn.
The generator templates are Python format strings, so literal JSON braces are escaped as \texttt{\{\{} and \texttt{\}\}} in the source files; the listings below are the messages after that escaping is resolved.
Arrows, en dashes, em dashes, and curly quotes in the source files are written here as \texttt{->}, \texttt{--}, and straight quotes.
The wording is otherwise unchanged.
Neither generator sends a system message: the listing is the entire user message.

\subsection{Follow-up query generator}
\label{sec:prompt-followup}

This prompt runs after the agent calls \texttt{finish}.
In the reported runs \texttt{\{recovery\_context\}} is empty, because those runs use the checker--fixer pipeline rather than the regenerate-from-scratch path that fills that slot, so the line below begins at ``Return''.
Temperature is 0.7.
The model must return only the JSON object in the prompt; \texttt{api\_calls} is an inspection hint and is not executed.

\begin{lstlisting}[style=prompt]
You are an AI assistant continuing a conversation with a user. The dialogue so far is provided between the <user_system_dialogues> tags, and the available tools you can invoke are described between the <apis> tags.

<user_system_dialogues>
{conversation_history}
</user_system_dialogues>

<apis>
{api_schema}
</apis>

Your task is to generate the NEXT user request in this conversation. This should be a natural follow-up that builds on the previous dialogue in a meaningful way.

**IMPORTANT: Generate a creative, insightful follow-up question that:**
1. Shows deeper engagement with the topic
2. Asks for related but different information than what's already discussed
3. Makes specific use of different parameters or capabilities
4. Demonstrates understanding of the previous context
5. Represents what a curious, thoughtful user might naturally ask next
6. Explores a new facet of the information already provided
7. **CRITICAL - ANTI-HALLUCINATION RULE: Only attribute facts to the assistant/tool/system if they were EXPLICITLY stated in the dialogue history. Do NOT claim "you said X" / "the tool returned X" unless it really happened.**
   - OK: If the assistant asked you to choose between options they provided (IDs, tickers, items, etc.), pick one explicitly.
   - OK: If you want to refine preferences (e.g., "show more from source X"), ensure X was explicitly present in the dialogue history.
   - NOT OK: Invent system/tool facts (IDs, records, exact names, numbers) that were not shown.

Remember that a real user would likely ask a specific, contextual follow-up rather than simply repeating their earlier question. They would reference information already provided and ask for something that builds on it. **However, a real user would only reference what was ACTUALLY said - they wouldn't make up details that weren't provided.**

GOAL / ENTITY ANCHORING (A) - OPTIONAL GUIDANCE:
- Generally keep the same core goal and entities from the dialogue history.
- Only re-anchor if there is a CLEAR and SEVERE drift to a completely wrong entity/topic. Do NOT over-correct minor shifts or create verbose re-anchoring statements.

CONTRADICTION RECONCILIATION (B) - OPTIONAL GUIDANCE:
- Only request verification if there are CLEAR and SEVERE conflicting facts about the same entity. Do NOT create loops by repeatedly asking for verification of minor inconsistencies.

Your follow-up should be solvable by invoking **exactly one** operation from <apis>, and should include specific parameters when appropriate.

Return **exactly** the JSON object below--no additional keys or text:
{
  "instruction": "<the creative, context-aware user follow-up you created>",
  "api_calls": [
    {
      "tool": "<tool name>",
      "operation": "<operation name>",
      "arguments": { <specific, relevant arguments based on the context> }
    }
  ]
}

**Constraints & Guidelines**
- Do **not** simply repeat or rephrase the previous user query
- The new instruction must logically follow from the preceding dialogue
- Do **not** invent capabilities beyond those listed in <apis>
- Be specific and precise about what information the user would want next
- **NEVER mention specific names, places, numbers, or details that the assistant did not explicitly provide in the dialogue. If you want to reference something, only reference what was ACTUALLY stated in the conversation history. Do NOT make up or infer specific details that weren't explicitly mentioned.**
- **Before referencing any specific information (like district names, document IDs, dates, locations, etc.) in your follow-up query, verify that this information was explicitly provided by the assistant in a previous message. If it wasn't, do NOT include it in your query.**
- Produce **valid JSON** with double quotes around all keys and string values
- Output absolutely nothing except the JSON object
\end{lstlisting}

\subsection{User-response generator}
\label{sec:prompt-user-response}

This prompt runs after the agent calls \texttt{message\_user}.
Temperature is 0.8.
\texttt{\{context\_section\}} is empty when the problem has no withheld context.
When it does, the slot expands to a blank line, the sentence ``Additional context about the user's preferences and situation:'', and a \texttt{<context>} block whose body is \texttt{Context:} followed by the withheld keywords (a comma-separated list, or \texttt{key: value} pairs).
The instruction is to weave those keywords into the reply rather than dump them.
As with the follow-up generator, \texttt{\{recovery\_context\}} is empty in the reported runs.

\begin{lstlisting}[style=prompt]
You are simulating a real user responding to an AI assistant in a conversation. The dialogue history is provided between the <dialogue_history> tags, and the available tools/APIs are described between the <apis> tags.{context_section}

<dialogue_history>
{conversation_history}
</dialogue_history>

<apis>
{api_schema}
</apis>

Your task is to generate a realistic user response to the assistant's last message. The assistant might be:
1. Asking for clarification about the user's original request
2. Requesting additional information to complete a task
3. Providing options for the user to choose from
4. Asking for confirmation before proceeding

**IMPORTANT: Generate a natural, helpful user response that:**
1. Directly addresses the assistant's question or request
2. Provides relevant and specific information when asked
3. Makes clear choices when presented with options
4. Expresses preferences or constraints in a natural way
5. **Uses the context information provided to inform your response** - incorporate relevant details from the context naturally
6. May include follow-up questions of their own
7. Uses conversational language (not overly formal or technical)
8. May occasionally express appreciation or frustration
9. **CRITICAL - ANTI-HALLUCINATION RULE: Only attribute facts to the assistant/tool/system if they were ACTUALLY stated in the dialogue history. Do NOT claim "you said X" / "the tool returned X" unless it really happened. If the assistant asks for information, respond to their request. If the assistant says they accessed something but doesn't provide details, do NOT assume you received those details.**

   **OK (and often necessary): Provide NEW user-provided details when asked (e.g., your location, a preference, a budget, an ID/value you want to use), as long as you do not present them as something the assistant/tool already said and you keep them consistent within the dialogue.**

   **NOT OK: Invent system/tool facts (IDs, records, prices, counts, tool outputs) that were not shown. If the assistant needs an ID that was not provided, ask them to list options/IDs so you can choose.**

10. **GOAL / ENTITY ANCHORING (A) - OPTIONAL GUIDANCE: Generally keep the same core goal and entities. If there is a CLEAR and SEVERE drift to a completely wrong entity/topic (e.g., asking about a Rolex when you want an Omega), you may gently correct it. However, do NOT over-correct minor topic shifts or create verbose re-anchoring statements that sound unnatural.**

11. **CONTRADICTION RECONCILIATION (B) - OPTIONAL GUIDANCE: If there are CLEAR and SEVERE conflicting facts about the same entity (e.g., dramatically different prices or specifications), you may ask for clarification. However, do NOT over-correct minor inconsistencies or create loops by repeatedly asking for verification.**

Remember that real users:
- Often provide more information than strictly necessary
- May reference their original intent/goal and context from their situation
- Sometimes change their mind or refine their request
- Express themselves in natural, conversational language
- Have varying levels of technical knowledge
- **Will naturally incorporate their context/preferences into their responses**

Return your response as a JSON object with the following format - don't include any other text:
{
  "user_response": "<the natural, realistic user response you created>"
}

**Guidelines:**
- Your response should be a single, natural-sounding message
- Be specific and detailed when providing information
- If the assistant has presented options, clearly indicate a preference based on your context
- If the assistant is confused, clarify your original intent using context information
- **When context is provided, naturally weave relevant details into your response**
- Occasionally include "human touches" like greetings, thanks, or mild impatience
- Focus on information relevant to the available APIs, but don't try to be overly technical
\end{lstlisting}

\subsection{Dialogue-quality judge}
\label{sec:prompt-quality}

Each aspect is one call.
The system message is \texttt{You are a dialogue evaluation expert focusing on <aspect>.}, with \texttt{<aspect>} equal to \texttt{consistency}, \texttt{relevance}, \texttt{efficiency}, \texttt{correctness}, or \texttt{adaptability}.
The user message is the template below.
The model must answer with a function call, \texttt{evaluate\_<aspect>}, whose arguments are the JSON object specified in that template.
The same five slots are filled in every template: \texttt{\{query\}}, \texttt{\{available\_tools\}}, \texttt{\{dialogue\}}, and \texttt{\{final\_answer\}}.
The weighted aggregate is computed from the five scores in code; there is no further judge call for the overall score.

\subsubsection{Consistency}

\begin{lstlisting}[style=prompt]
You are evaluating CONSISTENCY in tool usage for an AI assistant.

----------------------------------------
DEFINITION
----------------------------------------
Consistency measures whether the assistant maintains internally consistent use of tool-derived information across the conversation.

It evaluates whether the assistant:
- Reuses previously stated tool-based facts without changing them
- Avoids contradicting its own earlier statements
- Maintains stable references to values, entities, and results over time

----------------------------------------
IMPORTANT METRIC BOUNDARIES
----------------------------------------
- Do NOT evaluate whether the interpretation is correct (CORRECTNESS)
- Do NOT evaluate tool selection (RELEVANCE)
- Do NOT evaluate efficiency or number of calls (EFFICIENCY)
- Do NOT evaluate recovery from errors (ADAPTABILITY)

Even if the assistant initially misinterprets a tool result, it should NOT be penalized here as long as it remains consistent afterward.

----------------------------------------
EVALUATION STEPS (FOLLOW STRICTLY)
----------------------------------------
1. Identify key tool-derived facts used in the conversation (e.g., values, names, statuses).

2. Track how each fact is referenced across turns.

3. For EACH tracked fact:
   - Is it reused consistently?
   - Does any later statement contradict an earlier one?

4. You MUST classify all tracked references into:
   - stable_references
   - inconsistent_references

5. Grounding requirement:
   - Each inconsistency MUST include:
     - earlier statement
     - later contradictory statement

6. Special cases:
   - If no tool-derived facts are reused across turns:
     -> score = 1.0

----------------------------------------
CATEGORY DEFINITIONS
----------------------------------------
- stable_references:
  Facts that are reused consistently across the dialogue

- inconsistent_references:
  Facts that are changed, contradicted, or drift over time

----------------------------------------
SCORING RULES (DISCRETE)
----------------------------------------
- 1.0 -> No inconsistencies
- 0.75 -> Minor inconsistency with no material impact
- 0.5 -> Noticeable inconsistency affecting reliability
- 0.25 -> Major contradiction(s)
- 0.0 -> Repeated or severe contradictions

If unsure between two scores, choose the LOWER score.

----------------------------------------
OUTPUT FORMAT (STRICT JSON)
----------------------------------------
Return ONLY valid JSON:

{
  "stable_references": ["..."],
  "inconsistent_references": [
    {
      "earlier": "...",
      "later": "..."
    }
  ],
  "score": 0.0|0.25|0.5|0.75|1.0,
  "rationale": "Concise explanation focusing ONLY on internal consistency"
}

----------------------------------------
INPUT
----------------------------------------
Query: {query}

Available Tools:
{available_tools}

Dialogue Messages w/ Tool Calls:
{dialogue}


Final Response:
{final_answer}
\end{lstlisting}

\subsubsection{Relevance}

\begin{lstlisting}[style=prompt]
You are evaluating RELEVANCE in tool usage for an AI assistant.

----------------------------------------
DEFINITION
----------------------------------------
Relevance measures whether the assistant:
- Used tools only when needed
- Selected tools that matched the user's task
- Used parameters aligned with the request
- Made tool calls with a clear purpose toward solving the user's query

----------------------------------------
IMPORTANT METRIC BOUNDARIES
----------------------------------------
- Do NOT evaluate correctness of tool outputs (CORRECTNESS)
- Do NOT evaluate redundancy or number of calls (EFFICIENCY)
- Do NOT evaluate cross-turn consistency (CONSISTENCY)
- Do NOT evaluate recovery from failures (ADAPTABILITY)

Evaluate ONLY whether the chosen tools and their intended usage were appropriate for the user's task.

----------------------------------------
EVALUATION STEPS (FOLLOW STRICTLY)
----------------------------------------
1. Determine whether tools were needed at all for the task.

2. For EACH tool call:
   - Was the tool appropriate for the user's task?
   - Did the parameters align with the request?
   - Did the tool call have a clear purpose?
   - Important: the relevance of a tool call should be determined based on the relevance of the call and it's arguments, NOT the response. If the response of a call were not useful, this does not imply the call was unnecessary if it should or could have produced a useful result.

3. You MUST classify EVERY tool call into exactly ONE category:
   - appropriate_calls
   - questionable_calls
   - unnecessary_calls

4. If uncertain how to classify a tool call:
   -> classify it as questionable_calls

5. Special cases:
   - If no tools were needed and none were used:
     -> score = 1.0

   - If there are zero tool calls in the entire dialogue, but the query required tool use:
     -> score = 0.0

   - If tools were used when none were needed:
     -> classify them as unnecessary_calls



----------------------------------------
CATEGORY DEFINITIONS
----------------------------------------
- appropriate_calls:
  Tool calls that were clearly needed and appropriately aligned with the user's task, OR were clearly appropriate attempts at solving the user's tasks, but happened to produce errors or generic results.

- questionable_calls:
  Tool calls that were somewhat related but weakly justified, under-specified, uncertain, or only partially aligned with the task

- unnecessary_calls:
  Tool calls that were not needed or mismatched to the task

----------------------------------------
SCORING RULES (DISCRETE)
----------------------------------------
- 1.0 -> All calls appropriate, or no tools were needed and none were used
- 0.75 -> Mostly appropriate with minor questionable calls
- 0.5 -> Mixed appropriate and questionable/unnecessary calls
- 0.25 -> Several unnecessary or poorly aligned calls
- 0.0 -> Tool usage largely irrelevant or mismatched

----------------------------------------
OUTPUT FORMAT (STRICT JSON)
----------------------------------------
Return ONLY valid JSON:

{
  "needed_tools": true,
  "appropriate_calls": [],
  "questionable_calls": [],
  "unnecessary_calls": [],
  "score": 0.5,
  "rationale": ""
}

Rules:
- Replace needed_tools with true or false
- score MUST be one of: 0.0, 0.25, 0.5, 0.75, 1.0
- All arrays must always be present
- Never omit score
- Never output null for score
- Never output additional fields
- rationale must be concise
- Do NOT return explanations outside JSON

----------------------------------------
INPUT
----------------------------------------
Query: {query}

Available Tools:
{available_tools}

Dialogue Messages w/ Tool Calls:
{dialogue}


Final Response:
{final_answer}
\end{lstlisting}

\subsubsection{Efficiency}

\begin{lstlisting}[style=prompt]
You are evaluating EFFICIENCY in tool usage for an AI assistant.

----------------------------------------
DEFINITION
----------------------------------------
Efficiency measures whether the assistant reached the result using a MINIMAL and WELL-ORDERED sequence of tool calls.

Focus on:
- Avoiding redundant or repeated calls
- Avoiding unnecessary intermediate steps
- Using a logical sequence that minimizes total work

----------------------------------------
IMPORTANT METRIC BOUNDARIES
----------------------------------------
- Do NOT evaluate whether the correct tool was chosen (RELEVANCE)
- Do NOT evaluate correctness of outputs (CORRECTNESS)
- Do NOT evaluate consistency with tool results (CONSISTENCY)
- Do NOT evaluate recovery from failures (ADAPTABILITY)

Evaluate ONLY whether the number and sequence of calls were efficient.

----------------------------------------
EVALUATION STEPS (FOLLOW STRICTLY)
----------------------------------------
1. Determine the MINIMAL number of tool calls needed for this task (rough estimate).

2. For EACH tool call:
   - Did it provide NEW information or progress?
   - Was it redundant with previous calls?
   - Was it avoidable with better planning or sequencing?
   - If it was a retry, was it justified? (random or real-time APIs, possibly intermittent errors)
   - If it did not provide NEW information or progress, was the decision to call this API still justified?
   - Your decision should be informed by the tool call and arguments, not the response. You may consider the previous conversation and prior tool responses.

3. You MUST assign EVERY tool call to exactly ONE category:
   - efficient_calls
   - redundant_calls
   - justified_attempts
   - justified_retries

4. Special cases:
   - If no tools were needed:
     - If none were used -> score = 1.0
     - If tools were used -> they are redundant_calls

   - If no tools were ever used:
     - Score 1.0 even if tools should have been used.

   - If a call repeats but:
     - previous result failed, incomplete, or ambiguous
     - API is designed to produce random or real-time results that would likely change in the short duration of a conversation.
     -> classify as justified_retries (NOT redundant)


----------------------------------------
CATEGORY DEFINITIONS
----------------------------------------
- efficient_calls:
  Calls that contribute new information and are part of a minimal solution path

- redundant_calls:
  Calls that repeat previous work, add no new value, or could have been avoided

- justified_attempts:
  Calls that were well motivated, but happened to produce errors or non-useful results. A response to this call which could have contributed new information is plausible. These should not be penalized.

- justified_retries:
  Repeated or additional calls that are necessary due to failure, ambiguity, or missing data

----------------------------------------
SCORING RULES (DISCRETE)
----------------------------------------
- 1.0 -> Minimal number of calls, no redundancy
- 0.75 -> Minor inefficiency (1--2 slightly redundant or suboptimal steps)
- 0.5 -> Noticeable inefficiency (several redundant calls, but still arriving at a useful solution)
- 0.25 -> Major inefficiency (many redundant or unnecessary calls or poor sequencing)
- 0.0 -> Highly inefficient or chaotic tool usage

If unsure between two scores, choose the LOWER score.

----------------------------------------
OUTPUT FORMAT (STRICT JSON)
----------------------------------------
Return ONLY valid JSON:

{
  "minimal_calls_estimate": <integer>,
  "efficient_calls": ["..."],
  "redundant_calls": ["..."],
  "justified_retries": ["..."],
  "score": 0.0|0.25|0.5|0.75|1.0,
  "rationale": "Concise explanation focusing ONLY on redundancy and sequencing"
}

----------------------------------------
INPUT
----------------------------------------
Query: {query}

Available Tools:
{available_tools}

Dialogue Messages w/ Tool Calls:
{dialogue}


Final Response:
{final_answer}
\end{lstlisting}

\subsubsection{Correctness}

\begin{lstlisting}[style=prompt]
You are evaluating CORRECTNESS in tool usage for an AI assistant.

----------------------------------------
DEFINITION
----------------------------------------
Correctness measures whether the assistant ACCURATELY interpreted and used tool outputs.

Focus on:
- Faithfully reporting tool results
- Correctly interpreting tool success/failure
- Performing accurate calculations or transformations
- Avoiding unsupported claims that contradict tool outputs
- Detecting when the assistant invents factual claims after receiving non-actionable or documentation-like tool output

----------------------------------------
IMPORTANT METRIC BOUNDARIES
----------------------------------------
- Do NOT evaluate whether the correct tool was chosen (RELEVANCE)
- Do NOT evaluate number of calls or redundancy (EFFICIENCY)
- Do NOT evaluate cross-turn consistency (CONSISTENCY)
- Do NOT evaluate recovery behavior (ADAPTABILITY)

Evaluate ONLY whether tool outputs were USED CORRECTLY.

----------------------------------------
EVALUATION STEPS (FOLLOW STRICTLY)
----------------------------------------
1. Identify all tool outputs used by the assistant.

2. For EACH usage of tool output:
   - Was the information copied or summarized accurately?
   - Was the meaning interpreted correctly?
   - Were any calculations or transformations correct?

3. You MUST classify EACH usage into exactly ONE category:
   - correct_uses
   - minor_errors
   - material_errors

4. Grounding requirement:
   - Every error MUST reference a mismatch between:
     (a) tool output
     (b) assistant's usage in the dialogue or final answer

5. Special cases:
   - If no tools were used:
     - If the answer does not rely on tools -> score = 1.0
     - If the assistant claims tool-based facts without tools -> material error
   - If there are tool calls, but no assistant messages (zero in the entire dialogue):
     - Score 0.0

----------------------------------------
CATEGORY DEFINITIONS
----------------------------------------
- correct_uses:
  Tool outputs are used accurately and appropriately

- minor_errors:
  Small inaccuracies that do not significantly affect the final answer

- material_errors:
  Incorrect interpretation, wrong values, faulty calculations, or claims contradicting tool outputs

Additional material error cases:
- Claims presented as tool-derived facts when the tool returned only:
  - generic documentation
  - API descriptions
  - capability summaries
  - empty output
  - non-actionable information

----------------------------------------
SCORING RULES (DISCRETE)
----------------------------------------
- 1.0 -> All tool outputs used correctly
- 0.75 -> Minor errors with little impact
- 0.5 -> Noticeable errors affecting parts of the answer
- 0.25 -> Major errors affecting reliability
- 0.0 -> Severe or repeated misuse of tool outputs

If unsure between two scores, choose the LOWER score.

----------------------------------------
OUTPUT FORMAT (STRICT JSON)
----------------------------------------
Return ONLY valid JSON:

{
  "correct_uses": [],
  "minor_errors": [],
  "material_errors": [],
  "score": 0.5,
  "rationale": ""
}

Rules:
- score MUST be one of: 0.0, 0.25, 0.5, 0.75, 1.0
- All arrays must always be present
- Never omit score
- Never output null for score
- Never output additional fields

----------------------------------------
INPUT
----------------------------------------
Query: {query}

Available Tools:
{available_tools}

Dialogue Messages w/ Tool Calls:
{dialogue}


Final Response:
{final_answer}
\end{lstlisting}

\subsubsection{Adaptability}

\begin{lstlisting}[style=prompt]
You are evaluating ADAPTABILITY in tool usage for an AI assistant.

----------------------------------------
DEFINITION
----------------------------------------
Adaptability measures how the assistant responds when the original plan cannot proceed cleanly.

This includes reacting to:
- tool failures (errors, exceptions)
- incomplete or empty tool outputs
- ambiguous user requests requiring clarification
- unexpected results requiring strategy change
- unavailable tools requiring fallback behavior
- tool outputs that contain only generic documentation, API descriptions, capability summaries, or non-actionable information instead of usable task data

----------------------------------------
IMPORTANT METRIC BOUNDARIES
----------------------------------------
- Do NOT evaluate correctness of outputs (CORRECTNESS)
- Do NOT evaluate tool choice (RELEVANCE)
- Do NOT evaluate efficiency or number of calls (EFFICIENCY)
- Do NOT evaluate consistency across turns (CONSISTENCY)

Evaluate ONLY whether the assistant appropriately ADAPTED when required.

----------------------------------------
OPPORTUNITY DETECTION (STRICT)
----------------------------------------
Set opportunity_present = true ONLY if there is CLEAR evidence of at least one of:
- tool returned error / failure
- tool returned empty / incomplete / unusable result
- tool returned generic documentation, API descriptions, capability summaries, or non-actionable output instead of usable task data
- user query is ambiguous and requires clarification BEFORE tool use
- tool result contradicts expectations and requires change in approach

If NONE of the above occurred:
-> opportunity_present = false
-> score = 1.0

Do NOT invent hypothetical opportunities.

----------------------------------------
EVALUATION STEPS (FOLLOW STRICTLY)
----------------------------------------
1. Identify all adaptation-triggering events (if any).

2. For EACH event:
   - Did the assistant recognize the issue?
   - Did it respond appropriately (retry, clarify, switch strategy, fallback)?

3. Classify assistant behavior into:
   - appropriate_adaptations
   - missed_adaptations

4. Grounding requirement:
   - Each adaptation or miss MUST reference:
     - triggering event
     - assistant response

----------------------------------------
CATEGORY DEFINITIONS
----------------------------------------
- appropriate_adaptations:
  Clear, effective responses to a triggering event (retry, clarification, fallback, strategy change)

- missed_adaptations:
  Cases where the assistant failed to respond appropriately to a triggering event

Non-actionable outputs include:
- generic API descriptions
- documentation-like responses
- capability summaries
- responses without usable task-specific information

----------------------------------------
SCORING RULES (DISCRETE)
----------------------------------------
- If opportunity_present = false -> score = 1.0

Otherwise:
- 1.0 -> All events handled appropriately
- 0.75 -> Mostly handled well with minor misses
- 0.5 -> Some adaptation, but incomplete or weak handling
- 0.25 -> Poor handling of most events
- 0.0 -> Failed to adapt or made situation worse

If unsure between two scores, choose the LOWER score.

----------------------------------------
OUTPUT FORMAT (STRICT JSON)
----------------------------------------
Return ONLY valid JSON:

{
  "opportunity_present": true,
  "triggering_events": [],
  "appropriate_adaptations": [],
  "missed_adaptations": [],
  "score": 0.5,
  "rationale": ""
}

Rules:
- score MUST be one of: 0.0, 0.25, 0.5, 0.75, 1.0
- All arrays must always be present
- Never omit score
- Never output null for score
- Never output additional fields

----------------------------------------
INPUT
----------------------------------------
Query: {query}

Available Tools:
{available_tools}

Dialogue Messages w/ Tool Calls:
{dialogue}

Final Response:
{final_answer}
\end{lstlisting}

\subsection{Task-success judge}
\label{sec:prompt-task-success}

This prompt runs once per subtask, after the deterministic placeholder gate.
The system message is: \texttt{You judge end-to-end task success for tool-using assistants. Be strict on wrong or missing answers.}
The user message is below.
The model must answer with a function call \texttt{evaluate\_task\_success} whose arguments are \texttt{task\_succeeded} (boolean) and \texttt{justification} (one or two sentences).
\texttt{\{query\_kind\}} is \texttt{initial} or \texttt{follow\_up}.
\texttt{\{query\}}, \texttt{\{dialogue\}}, \texttt{\{tool\_calls\}}, and \texttt{\{final\_answer\}} are that subtask's slice only.

\begin{lstlisting}[style=prompt]
You judge whether the assistant **fully completed one specific user query** (outcome / goal satisfaction for that query only), not how polished the tool usage was.

This dialogue may contain multiple user goals across turns. Evaluate **only** the query below -- ignore whether other follow-up goals were met.

**Query under evaluation ({query_kind}):**
{query}

**Conversation for this query only (user/assistant turns and tool calls for this subtask):**
{dialogue}

**Tool call trace for this query (use to verify facts):**
{tool_calls}

**Final answer the user sees for this query:**
{final_answer}

Decide if this specific query's information need was fully met: correct target, no critical hallucination against tool results for this subtask, and a usable answer. Partial progress or success on other dialogue turns does not count. If the run clearly failed to answer this query (e.g. wrong tool domain, empty non-answer, max-turn forced incomplete answer), mark not succeeded.

Return structured output only via the function call: task_succeeded (boolean) and a short justification.
\end{lstlisting}

\end{document}